\documentclass[a4paper,fleqn]{cas-dc}

\usepackage[authoryear,longnamesfirst]{natbib}
\usepackage{balance}

\usepackage{amssymb}
\usepackage{soul}
\usepackage{xurl}
\usepackage{booktabs}
\usepackage{subcaption}
\usepackage{hyperref}
\hypersetup{hidelinks}
\usepackage{algorithm}
\usepackage{multirow}
\usepackage{makecell}
\usepackage[table]{xcolor}
\usepackage{etoolbox}
\usepackage{tikz}
\usepackage{enumitem}
\usepackage{amsmath,amsfonts}
\usepackage{array}
\usepackage{kotex}
\usepackage{seqsplit}
\usepackage{pifont}
\usepackage{algpseudocode}

\def\tsc#1{\csdef{#1}{\textsc{\lowercase{#1}}\xspace}}
\tsc{WGM}
\tsc{QE}

\begin{document}
\let\WriteBookmarks\relax
\def\floatpagepagefraction{1}
\def\textpagefraction{.001}

\newcommand{\ours}{\textsc{PassREfinder}}
\newcommand{\oursNew}{\textsc{PassREfinder-FL}}
\newcommand{\xmark}{\ding{55}}%

\newcommand{\squaretext}[1]{\raisebox{1.5pt}{\colorbox{#1}{\rule{0pt}{2pt}\rule{2pt}{0pt}}}}

\newcommand{\jh}[1]{{#1}}
\newcommand{\rev}[1]{{#1}}
\newcommand{\jw}[1]{{#1}}
\newcommand{\mk}[1]{{#1}}

\shorttitle{Privacy-Preserving Credential Stuffing Risk Prediction via Graph-Based Federated Learning of Cross-Website Password Reuse}    

\shortauthors{Kim \emph{et al.}}  

\title [mode = title]{\oursNew: Privacy-Preserving Credential Stuffing Risk Prediction via Graph-Based Federated Learning for Representing Password Reuse between Websites}  
\tnotemark[1] 

\tnotetext[1]{This work extends our previous study, \ours{}~\cite{kim2024passrefinder}, which was presented at IEEE S\&P 2024.}

\author[kaist]{Jaehan Kim}[orcid=0000-0001-8048-097X]
\ead{jaehan@kaist.ac.kr}
\author[kaist]{Minkyoo Song}[orcid=0009-0004-1597-2053]
\ead{minkyoo9@kaist.ac.kr}
\author[etri]{Minjae Seo}[orcid=0000-0001-9240-5213]
\ead{ms4060@etri.re.kr}
\author[kaist]{Youngjin Jin}[orcid=0009-0002-8992-8827]
\ead{ijinjin@kaist.ac.kr}
\author[kaist]{Seungwon Shin}[orcid=0000-0002-1077-5606]
\ead{claude@kaist.ac.kr}
\author[kwu]{Jinwoo Kim}[orcid=0000-0003-1303-8668]
\ead{jinwookim@kw.ac.kr}

\cormark[1]

\affiliation[kaist]{organization={KAIST},
addressline={291 Daehak-ro, Yuseong-gu}, 
city={Daejeon},
postcode={34141}, 
country={Republic of Korea}}

\affiliation[etri]{organization={ETRI},
    city={218 Gajeong-ro, Yuseong-gu},
    city={Daejeon},
    postcode={34129}, 
    country={Republic of Korea}
    }

\affiliation[kwu]{organization={Kwangwoon University},
            addressline={20 Kwangwoon-ro, Nowon-gu}, 
            city={Seoul},
            postcode={01897}, 
            country={Republic of Korea}}

\cortext[1]{Corresponding author}



\begin{abstract}
\jh{
Credential stuffing attacks have caused significant harm to online users who frequently reuse passwords across multiple websites. While prior research has attempted to detect users with reused passwords or identify malicious login attempts, existing methods often compromise usability by restricting password creation or website access, and their reliance on complex account-sharing mechanisms hinders real-world deployment. To address these limitations, we propose \oursNew{}, a novel framework that predicts credential stuffing risks across websites. We introduce the concept of password reuse relations---defined as the likelihood of users reusing passwords between websites---and represent them as edges in a website graph. Using graph neural networks (GNNs), we perform a link prediction task to assess credential reuse risk between sites. Our approach scales to a large number of arbitrary websites by incorporating public website information and linking newly observed websites as nodes in the graph. To preserve user privacy, we extend \oursNew{} with a federated learning (FL) approach that eliminates the need to share user sensitive information across administrators. Evaluation on a real-world dataset of 360 million breached accounts from 22,378 websites shows that \oursNew{} achieves an F1-score of 0.9153 in the FL setting. We further validate that our FL-based GNN achieves a 4-11\% performance improvement over other state-of-the-art GNN models through an ablation study. Finally, we demonstrate that the predicted results can be used to quantify password reuse likelihood as actionable risk scores.
}
\end{abstract}


\maketitle

\section{Introduction}

Credential stuffing---\textit{the use of a breached or stolen account from an online service to sign in to other online services}---is one of the most severe cyberattacks~\cite{verizon2022}. Due to a surge of massive credential data breaches, individual users and even various industries have suffered from the threat of credential stuffing. For example, retail industries have been imposed losses of 6 billion dollars per year~\cite{shape2018}. The devastating impact of credential stuffing has fundamentally resulted from the use of the same password by users across different online services. The problem of \textit{password reuse} has been repeatedly claimed by security researchers~\cite{das2014tangled, wash2016understanding, pearman2017let, thomas2017data, wang2018next}. Even though administrators have warned about the reuse of passwords to users, it was not sufficient to raise awareness for users to change their passwords~\cite{golla2018site}. \rev{Moreover, usability issues in current password managers remain, which can lead users to reuse the same passwords~\cite{hutchinson2024measuring}.}

To protect users from the risk of credential stuffing, several third-party services called compromised credential checking (C3) services~\cite{li2019protocols} have been developed, such as Have I Been Pwned (HIBP)~\cite{hibp}, Google Password Checkup~\cite{thomas2019protecting}. The C3 services alert users to reset passwords when the same passwords are reported from credential data breaches of other websites. Unfortunately, these services are limited to detecting the reuse of passwords disclosed by already known credential data breaches. As an alternative, a method of coordinating websites was proposed to detect the creation of the same password~\cite{wang2019end} or suspicious logins for credential stuffing attacks~\cite{wang2020detecting} by sharing usernames and passwords under privacy-preserving protocols. However, these approaches degrade the usability of passwords through direct interruption when choosing a password or denial of access to the websites due to the detection of false positives. More critically, the burdensome encryption methods preclude the deployment of their protocols at scale~\cite{pal2022might}, and the requirement of collecting usernames to map each username to websites discourages the participation of websites.

\jh{
As such, existing approaches face clear limitations, including reliance on known data, exposure of sensitive information, and limited scalability. To address these issues, we propose \oursNew{}, a privacy-preserving framework that predicts the risk of credential stuffing. Prior studies~\cite{pearman2017let, thomas2019protecting, wang2018next, gaw2006password, wash2016understanding, stobert2014password} have shown that website characteristics are closely linked to users’ password reuse behavior---for example, users often reuse passwords across websites in the same category or based on perceived security levels. Motivated by these findings, we analyze website features to capture their inherent influence on reuse behavior and predict the likelihood of credential stuffing between websites. To this end, we define a \textit{password reuse relation} as a relationship between websites where users are highly likely to reuse passwords, and our framework predicts the existence of such relations.
}

\jh{
To accomplish this goal, however, \emph{three} technical challenges must be addressed. First, password reuse relations across a large number of websites are inherently complex, as a user's password choice for one website is influenced by passwords used on others. To capture this, \oursNew{} models password reuse as edges in a graph of website nodes, leveraging graph neural networks (GNNs) to aggregate and propagate neighborhood influences. Second, the heterogeneous structure of website features and their varying impact on password reuse make them difficult to model with standard neural networks. To overcome this, \oursNew{} incorporates multi-modal learning~\cite{ngiam2011multimodal} and attention mechanisms~\cite{velivckovic2017graph, hori2017attention}, enabling the model to process each feature type independently and weigh their importance. Lastly, predicting relations involving unknown or new websites typically requires sharing password reuse information or user private information across potentially untrustworthy administrators, raising privacy concerns~\cite{kim2024passrefinder}. \oursNew{} mitigates this by employing a federated learning (FL) approach, allowing participants to share only model gradients or weights rather than sensitive information, thus preserving user privacy.
}

\jh{
We evaluate \oursNew{} through extensive experiments on a large-scale dataset comprising credential data breaches involving 360 million accounts across 22,378 websites. Our results demonstrate that the proposed GNN model under the FL setting effectively predicts password reuse relations between websites. In particular, it significantly outperforms our original framework, \ours{}~\cite{kim2024passrefinder} and its variants (i.e., with different hidden relation policies), achieving an F1-score of 0.9153. We also show that our GNN design yields performance improvements of up to 4-11\% compared to other state-of-the-art GNN models. We attribute the notable prediction performance of \oursNew{} to its FL-based approach, which effectively leverages the password reuse information across multiple administrators' websites. Furthermore, its privacy-preserving design enables administrators to confidently deploy the framework in real-world scenarios.
}

\jh{
In the ablation study, we assess the impact of each GNN modeling design choice in \oursNew{}. While the neighborhood attention method contributes less significantly under the FL setting---compared to its effectiveness in our original framework, \ours{}~\cite{kim2024passrefinder}, the multi-modality feature design consistently enhances prediction performance. Furthermore, we validate that the predicted probabilities directly reflect the expected rates of password reuse between websites, enabling their interpretation as credential stuffing risk scores. Overall, the results demonstrate that our model effectively learns edge representations for predicting password reuse relations across a large-scale graph of website nodes.
}

\noindent\textbf{Contributions.} Our contributions are summarized as follows:
\begin{itemize}
    \item We introduce the concept of password reuse relation to present the tendency of password reuse between websites and formalize the password reuse relation by leveraging a graph structure.
    \item We propose a new framework, \oursNew{}, to proactively predict the credential stuffing risk among websites based on graph neural networks, developing strategic techniques that can address technical challenges in the risk prediction task.
    \item \jh{We address the privacy concerns inherent in \ours{}, a credential stuffing risk prediction framework, by adopting an FL-based approach.}
    \item \jh{We evaluate our framework on a large-scale credential dataset, achieving an F1-score of 0.9153 and outperforming \ours{} while ensuring user privacy.}
    \item We show that our prediction results can be used to quantitatively measure credential stuffing risk scores among websites.
    \item We provide three practical application scenarios of \oursNew{} in the perspective of administrators to mitigate credential stuffing.
\end{itemize}

\section{Background}

In this section, we present the background knowledge and related studies.

\noindent\textbf{Credential stuffing detection.} Credential stuffing is recognized as a serious threat, and researchers have striven to prevent it by detecting password reuse. One of the most common techniques is a compromised credential checking (C3) service~\cite{thomas2019protecting, li2019protocols, pal2022might}, such as Have I Been Pwned (HIBP)~\cite{hibp}. C3 services provide breach-alerts when the same passwords are breached from other websites. However, existing C3 services can only detect reused passwords discovered from already disclosed data breaches. The discovery of data breach incidents took an average of 207 days, and containment lasted 70 days in 2022~\cite{ibm2022}. Hence, C3 services cannot provide timely coping with exploiting new accounts from undisclosed data breaches.

As another type of credential stuffing detection, Wang et al.~\cite{wang2019end} proposed a framework for detecting users who create the same passwords in different websites by coordinating the websites and sharing the information of usernames and passwords under a variant of private set-membership-test (PMT) protocol. Their following work~\cite{wang2020detecting} adopted the previous PMT protocol but modified its design for directly detecting login trials for credential stuffing attacks. Despite the promise of these approaches, they still have critical limitations. The detection frameworks extremely degrade the usability of passwords by immediately interrupting a user's password creation or denying website access due to false positives in detection. Moreover, the complicated encryption mechanisms make it difficult to deploy their protocols at scale~\cite{pal2022might}. Their username-website mapping system is also problematic because it requires all usernames from the websites, which discourages participation in the website coordination in practice.

\noindent\textbf{Password reuse analysis.} To prevent the devastating impact of credential stuffing, many studies on the risk of password reuse have been conducted. \rev{Specifically, analyzing features affecting password choice and reuse has been one of the main topics~\cite{alsabah2018your,mayer2017second, pearman2017let,thomas2019protecting, sahin2021don, wang2018next,gaw2006password,wash2016understanding,stobert2014password,moh2024understanding}.} For example, Alsabah et al.~\cite{alsabah2018your} and Mayer et al.~\cite{mayer2017second} claimed that users significantly considered their languages and websites' password composition policies to create their passwords. Pearman et al.~\cite{pearman2017let} presented a user study with 154 participants and discovered that the reuse of a password was influenced by its characteristics (e.g., password length and strength). In addition, they showed that the category of website performed a significant role in determining the reuse of passwords, as shown in several other works~\cite{thomas2019protecting,wang2018next,gaw2006password}. Wash et al.~\cite{wash2016understanding} warned that university students were likely to reuse their passwords in other university-related services. Stobert et al.~\cite{stobert2014password} showed that some users carefully consider the gap between the security aspects of two websites when they reused their passwords across websites. Moh et al.~\cite{moh2024understanding} found that a user often reuses the same password across personal and shared accounts, such as those shared with family members.

\noindent\textbf{Graph neural networks (GNNs).} Graph neural networks (GNNs) are neural networks for processing graph data, and have been widely utilized to represent relations between objects. As one of the most common types of GNN, graph convolutional neural network (GCN)~\cite{kipf2016semi} performs convolution operations in graphs based on message passing and neighborhood aggregation to represent the hidden features of nodes. GraphSAGE~\cite{hamilton2017inductive} is a variant of GCN for inductive representation of a large-sized graph by reflecting the self-embedding separately and sampling the neighborhood for aggregation. GAT~\cite{velivckovic2017graph} adopts a self-attentional aggregation approach to specify different weights to different nodes. GNN variants have achieved notable performance for various learning tasks in a wide range of security research domains~\cite{liu2021pick, yang2022wtagraph,cui2022meta,kim2023drainclog,seo2024gshock,lee2024censor}.

\noindent\jh{\textbf{Federated learning (FL) on graph.} Federated learning (FL) has recently emerged as a promising paradigm for training neural networks on decentralized data, addressing privacy concerns in the collaborative development of deep learning models across multiple entities~\cite{konevcny2016federatedlearning,konevcny2016federatedoptimization}.
In graph-based FL~\cite{he2021fedgraphnn}, multiple clients independently train GNNs using their local subgraphs. Each client computes local updates and transmits the resulting gradients or model weights to a central server or neighboring clients. These updates are then aggregated to refine a global GNN model, subsequently redistributed to the clients, thus preserving data privacy by avoiding direct sharing of raw graph data.
Scardapane et al.~\cite{scardapane2020distributed} introduce a fully distributed framework for training GCNs, enabling distributed inference and optimization across clients communicating over a sparse network without a central server.
Wu et al.~\cite{wu2021fedgnn} propose FedGNN, an FL framework specifically designed for recommendation systems, incorporating local differential privacy and pseudo-interaction sampling to securely model user-item interactions.
Recent research efforts have focused on improving communication efficiency within federated GNN training frameworks, aiming to reduce communication overhead and enhance scalability~\cite{yao2023fedgcn,zhang2021subgraph}.}
Recently, graph federated learning has been widely applied to security domains such as anomaly detection~\cite{kong2024federated,mao2025fecograph}, fraud detection~\cite{tang2024credit}, and malware detection~\cite{amjath2025graph}. In contrast, \oursNew{} introduces a novel problem formulation for the application of graph federated learning to protect user credentials.

\section{Motivation}
Here, we motivate the need for our work by introducing our previous study~\cite{kim2024passrefinder}, which highlights the limitations of existing approaches and lays the foundation for our proposed solution.

\begin{figure}
    \centering
    \includegraphics[width=\linewidth]{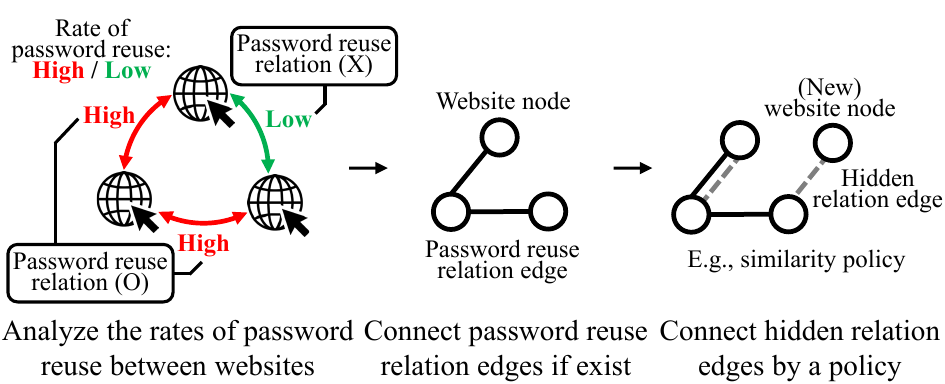}
    \caption{Graph representation of password reuse relations.}
    \label{fig:motivation}
\end{figure}

\noindent\textbf{Credential stuffing risk prediction.} As previously discussed, detecting credential stuffing is a reactive measure that fails to prevent attacks before they occur. A more proactive approach is to predict a website’s risk level by learning patterns such as website-specific characteristics and password reuse behaviors across websites. \ours{}~\cite{kim2024passrefinder} introduces a state-of-the-art method in this domain, leveraging GNNs to model users’ password reuse tendency. It frames the task as a link prediction problem---representing password reuse as a \emph{password reuse relation} and estimating whether a password used on one website is likely to be reused on another, as illustrated in Figure~\ref{fig:motivation}. However, directly exchanging password reuse information for its training or inference process raises significant ethical and privacy concerns. To address this, \ours{} introduced \emph{hidden relation}, a complementary edge connected by a pre-defined policy, propagating implicit information about password reuse tendencies to effectively learn credential stuffing risks across websites managed by different administrators. This allows administrators to \emph{proactively} identify potential vulnerabilities to credential stuffing, rather than responding only after an attack has occurred.

\noindent\textbf{Limitations of existing prediction methodology.} \ours{} has two key limitations: i) its sub-optimal design relies on heuristically learning implicit information through hidden relations, rather than leveraging ground truth password reuse information from other administrators; and ii) it poses inherent privacy risks, as it requires access to users’ private information (e.g., overlapping usernames) across potentially untrustworthy administrators to achieve optimal performance. These critical limitations hinder the practical deployment of \ours{} in real-world settings involving untrusted parties. This highlights the need for a more advanced approach that both preserves user privacy and effectively predicts credential stuffing risks.

\noindent\textbf{Our approach: \oursNew{}.} Our goal is to predict password reuse relations between websites without compromising user privacy. To this end, we extend \ours{} by removing the sub-optimal and privacy-sensitive hidden relation method and incorporating federated learning (FL), which mitigates the privacy risks associated with sharing password reuse data across websites. Instead of exchanging raw data, administrators share only the gradients or weights of their local GNNs, which are then used to collaboratively train a global model and construct a \emph{cross-admin password reuse relation}. Thanks to the privacy-preserving nature of FL, these shared gradients or weights do not disclose any sensitive information. In this way, our approach enables secure and privacy-preserving prediction of credential stuffing risks.

\section{\oursNew{} Overview}

\jh{The goal of \oursNew{} is to predict the existence of a password reuse relation---a relation between websites where users are highly likely to reuse passwords. \rev{Consistent with our original work, \ours{}~\cite{kim2024passrefinder}, we formally define that a password reuse relation exists between two given websites if the rate of \textit{users who reuse passwords} to \textit{users registered in both websites} (i.e., password reuse rate) is larger than the ground truth threshold $\tau$}\footnote{\rev{Although prior measurement studies have commonly estimated credential stuffing risk using the values of password reuse rates or password strengths~\cite{das2014tangled,pearman2017let,thomas2017data,hanamsagar2018leveraging,wang2023no,wang2016fuzzypsm}, we require a binary criterion to classify risk as high or low for a concrete and intuitive evaluation of our method. Accordingly, we introduce a tunable ground truth threshold $\tau$, consistent with \ours{} (our prior work), for balancing a trade-off between prediction performance and granularity}.}. To evaluate our novel approach based on graph federated learning, we particularly focus on predicting cross-admin password reuse relations, which are relations among websites managed by different administrators.}

\subsection{Challenges and Approaches}

We now discuss three key challenges in achieving the aforementioned goal, along with our approaches to address them.

\noindent\textbf{Complicated password reuse relations (C1).} The password reuse relations among multiple websites are highly complicated to be represented formally because users' password choice is determined by the passwords already used in other websites.
Thus, we need to analyze thousands of websites and their mutual influences in an attempt to predict password reuse relations at the website level. An intuitive method is to utilize the similarity between websites~\cite{pearman2017let, wash2016understanding}. However, in our evaluation, we present that password reuse relations are not able to be represented solely by the similarity of website features due to their implicit nature. Even existing prediction models such as machine learning classifiers have difficulty representing the complicated relations of password reuse among a large number of websites. 

To address this challenge, we leverage a graph structure that can represent edges for password reuse relations among website nodes. Then, we predict the existence of a password reuse relation edge between two given website nodes based on graph neural networks (GNNs), which are dedicated to learning complicated relations by benefiting from the propagation and aggregation of neighborhood influences.

\noindent\textbf{Diverse data structures and different influences of features (C2).} According to previous empirical studies, the extent of password reuse among websites is affected by diverse types of website features~\cite{pearman2017let, thomas2019protecting, wang2018next, gaw2006password, wash2016understanding, stobert2014password}. For example, users can sign in to online news websites with general passwords, but they use unique passwords for financial websites, and they tend to use different passwords for websites that have different security levels. In other words, it is challenging to comprehensively consider the website features for the task of predicting password reuse relations. The first reason is that the data structures of the website features are entirely different from each other (e.g., location, URL, and text). This makes it difficult for neural networks to learn the representations and leads to performance degradation in the prediction (see Section~\ref{subsec:ablation_study}). Moreover, the influence of a feature differs not only by the feature but also by specific website relationships. 

Therefore, defining which is more important for prediction is complicated. Thus, we adopt the technique of \textit{multi-modal features}~\cite{ngiam2011multimodal} and attention mechanisms~\cite{velivckovic2017graph, hori2017attention} to learn the diverse data structures of the website features separately and reflect their different quantitative importance.

\noindent\textbf{Difficulty in applying a graph-based model to arbitrary websites (C3).} \jh{In practice, website administrators may seek to predict the password reuse relation between a website they manage and one managed by another administrator, referred to as a \emph{cross-admin password reuse relation}. However, predicting these relations is challenging due to the limited access to password reuse information from external websites. To address this, \ours{}~\cite{kim2024passrefinder} introduces the \textit{hidden relation}, which links external website nodes using complementary edges rather than password reuse relation edges. Despite its effectiveness, the original design of \ours{} has two critical limitations. First, the method is inherently sub-optimal as it establishes hidden relations heuristically rather than based on ground truth (i.e., password reuse information) from external websites, which is crucial for accurately predicting cross-admin password reuse relations. Second, it requires private user information (e.g., matched users across websites) to optimize prediction performance. This potential privacy risk may discourage website administrators from jointly deploying \ours{} although broad participation is essential for maximizing its accuracy and coverage.
}

\jh{
To overcome these limitations, we propose a graph federated learning approach to predict password reuse relations in a privacy-preserving manner. This enhancement allows \oursNew{} to independently learn the password reuse relations associated with each administrator's websites without sharing sensitive information, such as passwords and usernames, across different administrators.
}

\begin{figure*}
    \centering
    \includegraphics[width=\linewidth]{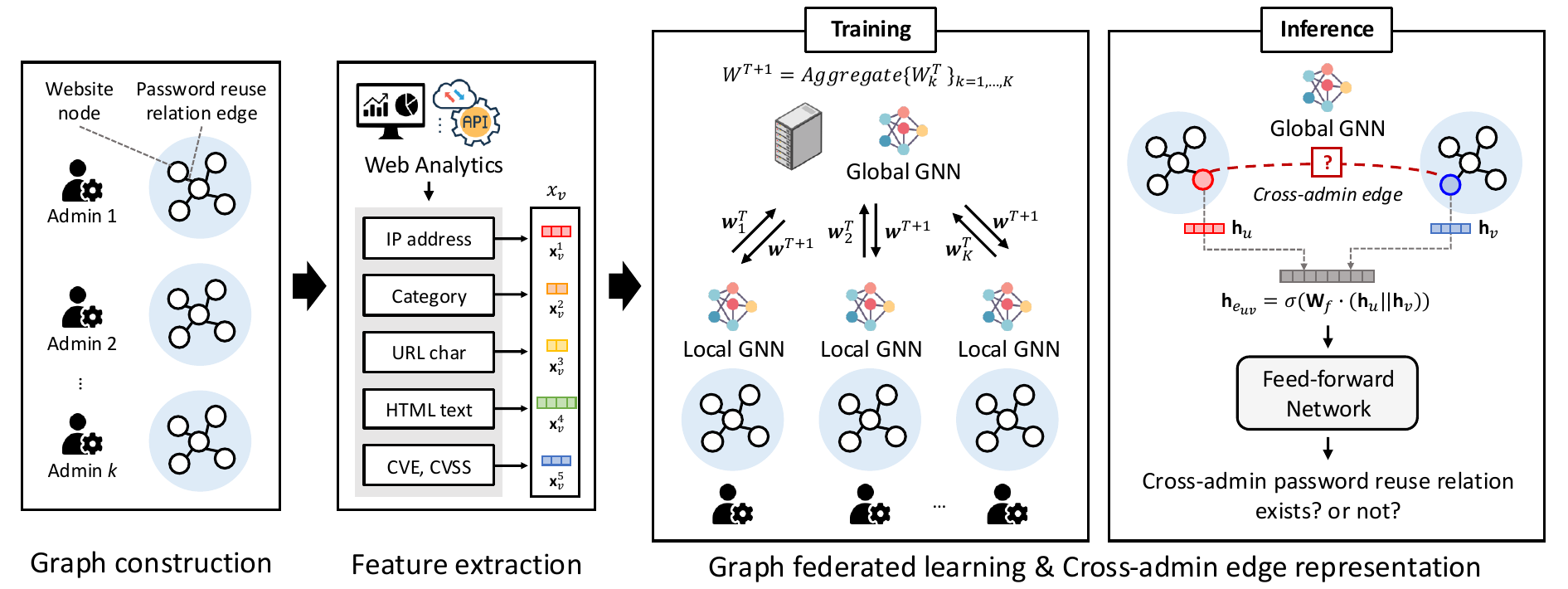}
    \caption{\jh{Overview of \oursNew{}.}}
    \label{fig:overview}
\end{figure*}

\subsection{Overall Architecture}

\jh{We design each component of \oursNew{} (see Figure~\ref{fig:overview}) to address the aforementioned challenges through a series of strategic design choices. Specifically, we aim to derive effective edge representations from a password reuse graph in a privacy-preserving manner. In contrast to the original work~\cite{kim2024passrefinder}, our enhanced framework is explicitly designed for deployment across multiple administrators who do not necessarily trust each other. Specifically, \oursNew{} incorporates three components:}

\noindent\textbf{Graph construction.} \jh{\oursNew{} first builds a password reuse graph using websites under that administrator's management. Two website nodes are connected by a password reuse relation edge if a password reuse relation exists between them. We note that administrators do not need to share any information about their password reuse graphs with one another. New administrators can integrate into our framework alongside existing administrators by constructing password reuse graphs from their own websites.}

\noindent\textbf{Feature extraction.} Next, \oursNew{} extracts various website features that can affect user password reuse tendencies. The features of a given website are automatically extracted by querying the information of the website from several public sources of web analytics~\cite{urlscan, mcafee, shodan, cachrome, camac, camozilla}. The method of feature extraction utilizing public information of websites enables our framework to be applied to arbitrary websites on the Internet. To reflect various types of website features, we adopt an embedding method dedicated to vectorizing each data structure as an effective input for a neural network. In addition, we establish the website features in multiple modalities to interpret different types of features separately.

\noindent\textbf{Graph federated learning \& Cross-admin edge representation.} \jh{\oursNew{} adopts a federated learning approach to train a GNN-based model tailored to producing representations of password reuse relation edges for predicting their existences. The detailed architecture of the GNN model follows the original design~\cite{kim2024passrefinder}. During training, each administrator updates the weights of the corresponding local GNN based on the local password reuse graph and then sends it to the central server for storing the global GNN. These updated weights are then sent to a central server, where they are aggregated to form the global GNN model. The server subsequently distributes the aggregated global GNN weights back to all administrators for the next training step.
}

\jh{
During inference, each administrator computes the node representation on the local password reuse graph and the global GNN weights they have received. To make a prediction on a cross-admin password reuse relation edge, the involved administrators exchange only their respective node representation vectors without sharing any sensitive information such as passwords and usernames. The edge representation vector then is constructed by concatenating the two exchanged node representation vectors. Finally, we predict the existence of a cross-admin password reuse relation edge by processing this edge representation vector using a feed-forward network.}

\section{\oursNew{} Design}
\label{sec:design}

This section details the design of \oursNew{}.

\subsection{Graph Construction}
\label{subsubsec:graph_construction}

\jh{Password reuse relations can be represented on a graph of websites. We formulate a \textit{password reuse graph} $\mathcal{G = (V, E)}$ to model the existence of password reuse relations among websites. We denote a set of website nodes as $\mathcal{V}$ and a set of password reuse relation edges as $\mathcal{E}$. Website nodes $u, v \in \mathcal{V}$ are connected by a password reuse relation edge $e_{uv} \in \mathcal{E}$ if a password reuse relation exists between $u$ and $v$. Note that a password reuse graph is undirected and has no self-loops because a password reuse relation can be defined for an unordered pair of two websites.}

\jh{In \oursNew{}, each administrator constructs a local password reuse graph based on (partial) information about password reuse relations among the websites the administrator manages. Specifically, the graph consists of website nodes connected by edges representing password reuse relations. An edge is established between two website nodes if the proportion of users reusing their passwords across those sites exceeds the ground truth threshold $\tau$. These local password reuse graphs are not interconnected---there are no password reuse relation edges across websites managed by different administrators. These cross-admin password reuse relation edges are the prediction targets. Unlike the original design of \ours{}~\cite{kim2024passrefinder}, which incorporates hidden relations, our approach ensures that no privacy-sensitive data and even the local password reuse graph information are shared across administrators.}

\jh{An administrator who deploys the framework uses their own constructed password reuse graph to train the GNN model, which limits the model’s ability to generalize to password reuse graph patterns of websites managed by other administrators. In contrast, \oursNew{} can leverage a broader range of password reuse patterns across all participating administrators by learning from distributed information of their local password reuse graphs.}

\subsection{Feature Extraction}
\label{subsubsec:feature_extraction}

Website node $v \in \mathcal{V}$ has an input feature (set) $x_v$ which is established from the website's characteristics. There are diverse website features affecting users' tendency to reuse their passwords, and the data structures of such features are different from each other. Therefore, we extract features based on \textit{$M$ modalities}, each of which denotes a single type of website feature. We let $\textbf{x}_v^m \in \mathbb{R}^{D_m}$ be the $m$-th feature vector with dimension $D_m$ and $x_v = \{\textbf{x}_v^1, \textbf{x}_v^2, ..., \textbf{x}_v^M\}$ be a feature set of website node $v$\footnote{In our prototype, we select website features of five modalities (i.e., $M = 5$), including the \textit{location}, \textit{category}, \textit{content}, \textit{URL}, and \textit{security posture} of a website.}. The features of website nodes are extracted from website characteristics that affect password choice and reuse. We note that the website features are obtained from the public services of web analytics. In our prototype, we establish five modalities of website features with different data structures to effectively represent password reuse relations. In what follows, we describe each of the features, the source of information, and the embedding method for creating input for the GNN layers.

\textbf{Location} is an effective website feature for understanding the reuse of passwords. For example, user password selection habits (e.g., password strength) significantly differ by country due to specific cultures of password composition policies and languages used to create passwords~\cite{mayer2017second, alsabah2018your}. Then, users may decide to reuse passwords according to the location where websites are serviced. We identify the locations of websites using their IP addresses retrieved from \textit{urlscan.io}~\cite{urlscan}, a popular website scanning service. To consider subnets in a more fine-grained manner, we encode the IP addresses into 32-bit binary embedding vectors.

\textbf{Category} is one of the well-known features correlated to password reuse. Users typically prioritize their accounts by the importance of the website~\cite{thomas2019protecting, wang2018next, pearman2017let, gaw2006password}. In particular, websites related to financial and governmental services show that users highly avoid reusing passwords in other websites due to high concerns about the exfiltration of private or sensitive information. On the other hand, users are more likely to reuse the passwords of unimportant websites (e.g., video streaming and disposable account websites). For the classification, we utilize the categorization service of McAfee~\cite{mcafee}. Then, we merge a few categories to mitigate the inconsistencies in the levels of some categories and finally limit the results to 20 categories. The category of a website cannot be represented by a numerical value or a vector. Therefore, we utilize the method of lookup embedding to obtain learnable vectors, which has been widely adopted to produce dense representations of (categorical) data~\cite{he2017neural, tang2018personalized}.

\textbf{Content} complements the category feature as website category is sometimes ambiguous~\cite{sherman2020} (e.g., providing multiple services). For the content feature, the HTML text of website pages can be used to extract information explicitly related to their services. Nevertheless, we use both the category and content features as some websites contain few texts on their pages. Notably, the text content can effectively reflect the languages of websites and users. In this regard, we use a transformer-based multilingual language model, XLM-RoBERTa~\cite{conneau2019unsupervised}, which is a powerful model for tasks in various domains~\cite{das2021heuristic, bozkir2022grambeddings, khatua2022unraveling}. We utilize the embedding vectors obtained from the pre-trained model~\cite{xlm-roberta-base}.

\textbf{URL} is a significant indicator of the relationship between two websites, such as the root domain and subdomain. If the URLs of two websites are related to each other, users tend to use the same usernames and passwords in the websites. For example, university students reuse their passwords in other university services that have similar URLs to that of the university homepage (i.e., the root domain)~\cite{wash2016understanding}. We adopt a method of character-level embedding for URL using Long Short-Term Memory (LSTM), modified from an existing URL embedding approach~\cite{yang2019phishing}.

\textbf{Security posture} is an important feature that users consider when reusing their passwords because it is directly related to the level of trust on a website~\cite{gaw2006password, stobert2014password}. We conduct a security analysis on website systems through \textit{Shodan}~\cite{shodan}, a search engine for Internet devices, and extract software vulnerabilities in a Common Vulnerabilities and Exposures (CVE)~\cite{cvesite} format. We first select the number of software operating in a website system as a large number of software extends the attack surface. We also utilize the average number of CVEs in each software and the average/maximum values of Common Vulnerability Scoring System (CVSS) scores~\cite{cvss} to measure the severities of the vulnerabilities. In addition, we examine HTTPS adoption, a more conspicuous security posture for users by validating website certificate chains. We use \textit{OpenSSL} by referring to the trusted certificate authorities of Google Chrome~\cite{cachrome}, Mozilla Firefox~\cite{camozilla}, and Apple macOS~\cite{camac} to analyze certificate errors such as certificate expiration, self-signed certificate (chain), unverifiable first certificate in chain and unverifiable local issuer, which is similar to the existing approach~\cite{singanamalla2020accept}. 
To stabilize the neural network training process while managing diverse scales of security posture features, we adopt batch normalization techniques~\cite{ioffe2015batch}.

\subsection{\jh{Graph Federated Learning \& Cross-Admin Edge Representation}}

\jh{We follow the design of GNNs introduced in our previous work, \ours{}~\cite{kim2024passrefinder}, which demonstrates its tailored capability in representing password reuse relations. Specifically, we adopt the novel aggregation function and two different levels of attention mechanisms, which enable us to model various website features and edges related to users' password reuse tendency. Subsequently, we introduce a federated learning algorithm for representing password reuse relations based on local password reuse graphs. We note that the hidden relation method of \ours{} has been entirely removed in \oursNew{}.}

\jh{
Formally, we consider a central server and $K$ website administrators as clients. Each administrator $k \in \{{1, ..., K}\}$ builds a password reuse graph $\mathcal{G}_k = (\mathcal{V}_k, \mathcal{E}_k)$ based on the set $\mathcal{V}_k$ of websites that the administrator manages and the set $\mathcal{E}_k$ of password reuse relations among these websites. Initially, the password reuse graph of an administrator is disconnected with other administrators' graphs. Now, we want to predict the existence of a cross-admin password reuse relation edge $e_{uv}$, where $u \in V_k, v \in V_{k'}$ ($k \neq k'$). To this end, the node representation of website node $u$ is obtained by applying GNNs on the password reuse graph $\mathcal{G}_k$ with website features $\{x_w: w \in \mathcal{V}_k\}$. Independently, the node representation of website node $v$ is obtained from the password graph $\mathcal{G}_{k'}$ with website features $\{x_w: w \in \mathcal{V}_{k'}\}$. Based on the node representations of $u$ and $v$, our model generates $d$-dimensional password reuse relation edge representations $\{\textbf{h}_e \in \mathbb{R}^d: e \in \mathcal{E}\}$ for facilitating the link prediction task regarding (cross-admin) password reuse relations.
}

\begin{algorithm}
    \renewcommand{\algorithmicrequire}{\textbf{Input:}}
    \renewcommand{\algorithmicensure}{\textbf{Output:}}
    \caption{\jh{Local node representation learning}}
    \begin{algorithmic}[1]
        \Require Local password reuse graph $\mathcal{G = (V, E)}$;
        the number of modalities $M$;
        the number of GNN layers $L$;
        input node feature set $x_v = \{\textbf{x}_v^1, \textbf{x}_v^2, ..., \textbf{x}_v^M\}$;
        weight matrices $\textbf{W}^{m, l}, \forall m \in \{1, ..., M\}, \forall l \in \{1, ..., L\}$;
        activation function $\sigma$;
        neighborhood functions $\mathcal{N}: v \rightarrow 2^{\mathcal{V}}$;
        neighbor aggregation function $\textsc{Agg}$;
        modality attention function $\textsc{ModalityAttn}$
        \Ensure  vector representations $\textbf{h}_{v}, \forall v \in \mathcal{V}$
        \For{$m = 1 ... M$}
            \State $\textbf{h}^{m, 0}_{v} \gets \textbf{x}_v^m, \forall v \in \mathcal{V}$
            \For{$l = 1  ... L$}
                \For{$v \in \mathcal{V}$}
                    \State $\textbf{h}^{m, l}_{\mathcal{N}(v)} \gets \textsc{Agg} (\textbf{h}^{m, l - 1}_{u}, \forall u \in \mathcal{N}(v))$
                    \State $\textbf{h}^{m, l}_{v} \gets \sigma (\textbf{W}^{m, l} \cdot \textsc{Concat} (\textbf{h}^{m, l - 1}_{v}, \textbf{h}^{m, l}_{\mathcal{N}(v)}))$
                \EndFor
            \EndFor
            \State $\textbf{h}^{m}_{v} \gets \textbf{h}^{m, L}_{v} / \lVert \textbf{h}^{m, L}_{v} \rVert_2, \forall v \in \mathcal{V}$
        \EndFor
\State $\textbf{h}_{v} \gets$ $\textsc{ModalityAttn} (\textbf{h}^{m}_{v}, \forall m \in \{1, ..., M\}), \forall v \in \mathcal{V}$
    \end{algorithmic}
    \label{alg:learning}
\end{algorithm}

\jh{
We now describe how to learn vector representations for website nodes in each administrator's local password reuse graph, using website-specific features. The overall procedure is illustrated in Algorithm~\ref{alg:learning}. The main inputs to the algorithm include a local password reuse graph $\mathcal{G}$ and, for each node, a feature set $x_v$ spanning $M$ modalities. The number of GNN layers is denoted by $L$. For each modality and layer, the algorithm uses a total of $M \times L$ uniformly initialized weight matrices $\mathbf{W}^{m, l}$. The algorithm further relies on a neighborhood function $\mathcal{N}$, which returns the neighbors of a given node in $\mathcal{G}$. Lastly, it takes as input the neighbor aggregation and modality attention functions we defined. The goal is to obtain a vector representation $\mathbf{h}_v$ for each website node in the graph $\mathcal{G}$.
}

\noindent\textbf{Feature initialization.} The algorithm begins by initializing the node representation vectors (lines 1 to 2). For each modality $m$, we assign the input feature $\textbf{x}_{v}^{m}$ to the node representation vector of the initial layer $\textbf{h}_{v}^{m, 0}$.

\noindent\textbf{Neighbor aggregation.} A website node's representation is shaped not only by its own features, but also by the features of its neighbors. Thus, for each GNN layer $l$ (and for each modality $m$), the node $v$'s representation vector is produced by aggregating that of its neighbors (lines 4 to 7). Aggregated neighbors are determined by the neighborhood function $\mathcal{N}$. To prioritize neighbors, we adopt neighborhood attention during the aggregation. It takes the node representation vector $\textbf{h}_{v}^{m, l - 1}$ and each neighbor $u$'s node representation vector $\textbf{h}_{u}^{m, l - 1}$ as inputs to produce the attention coefficient $\alpha_{v}^{m, l} (u, t)$ that denotes the influence of a neighbor node $u$ on the given node $v$ propagated through edges. The attention coefficient is formulated as follows:
\begin{align}
    \widetilde\alpha_{v}^{m, l} (u) = \sigma (\textbf{a}_{m, l}^{T} (\textbf{W}^{m, l} \textbf{h}_{u}^{m, l - 1} + \textbf{W}^{m, l} \textbf{h}_{v}^{m, l - 1})),
\end{align}
\noindent{}where $\textbf{W}^{m, 1} \in \mathbb{R}^{D_{m} \times d}$ and $\textbf{W}^{m, l( > 1)} \in \mathbb{R}^{d \times d}$ denote weight matrices for the node representation vectors, $\textbf{a}_{m, l} \in \mathbb{R}^{d}$ denotes a weight vector for parameterizing a single scalar value, and $\cdot^T$ is the transposition operator.

Then, we normalize the attention coefficients across all neighbor nodes using the softmax function:
\begin{align}
    \alpha_{v}^{m, l} (u) = \frac{\text{exp} (\widetilde\alpha_{v}^{m, l} (u)) }{\Sigma_{w \in \mathcal{N}(v)} \text{exp} (\widetilde\alpha_{v}^{m, l} (w))}.
\end{align}
\noindent{}We can obtain the aggregated neighborhood vector using a linear combination of the neighbors' node representation vectors based on the normalized attention coefficients:
\begin{align}
    \textbf{h}^{m, l}_{\mathcal{N}(v)} = \sum_{u \in \mathcal{N}(v)} \alpha_{v}^{m, l} (u) \textbf{h}_{u}^{m, l - 1}.
\end{align}
\noindent{}After the aggregation on each node, we impose different weights to the node representation vector $\textbf{h}_{v}^{m, l - 1}$ and the aggregated neighborhood vector $\textbf{h}^{m, l}_{\mathcal{N}(v)}$ by concatenating the two vectors and multiplying its result by the weight matrix $\textbf{W}^{m, l}$, which is effective for inductive learning~\cite{hamilton2017inductive} (\textit{cf}. adding them in GCN~\cite{kipf2016semi}). Subsequently, we obtain node $v$'s representation vector $\textbf{h}_{v}^{m, l}$ in layer $l$.

\noindent\textbf{Multi-modality attention.} After we normalize the node representation vectors of the last layer $L$ (line 9), we apply the modality attention function across the node representation vectors of $M$ modalities (line 11). The normalized modality attention coefficients are formalized as follows:
\begin{align}
    \beta_{v} (m) = \frac{\text{exp} (\sigma (\textbf{b}_{m}^{T} \textbf{W}_{1} \textbf{h}_{v}^{m} + b)) }{\Sigma_{m' = 1}^{M} \text{exp} (\sigma (\textbf{b}_{m'}^{T} \textbf{W}_{1} \textbf{h}_{v}^{m'} + b))},
\end{align}
\noindent{}where $\textbf{W}_{1} \in \mathbb{R}^{d \times d}$ denotes a weight matrix for the node representations of the modalities, $\textbf{b}_{m} \in \mathbb{R}^{d}$ denotes a weight vector for parameterizing a single scalar, and $b$ denotes a scalar bias. The node representation vector $\textbf{h}_{v}$ is obtained through a linear combination of the modalities’ node representation vectors multiplied by the normalized modality attention coefficients:
\begin{align}
    \textbf{h}_{v} = \sum_{m = 1}^{M} \beta_{v} (m) \textbf{h}_{v}^{m}.
\end{align}

\noindent\jh{\textbf{Edge representation under the FL setting.} To infer password reuse relationships between websites managed by different administrators---without exposing sensitive user data---we extend our original work~\cite{kim2024passrefinder} with a federated learning (FL) approach. In this setting, each administrator maintains a local password reuse graph $\mathcal{G}_k = (\mathcal{V}_k, \mathcal{E}_k), \forall k \in \{1, ..., K\}$ and we aim to collaboratively train a GNN model across all $K$ administrators. Our objective is to learn an edge representation vector $\mathbf{h}_{e_{uv}}$ for each pair of websites $u \in \mathcal{V}_k$ and $v \in \mathcal{V}_{k'}$ belonging to different administrators $k \ne k'$.}

\begin{algorithm}
    \renewcommand{\algorithmicrequire}{\rev{\textbf{Input:}}}
    \renewcommand{\algorithmicensure}{\rev{\textbf{Output:}}}
    \caption{\rev{Graph federated learning}}
    \begin{algorithmic}[1]
        \Require \rev{Central server $S$; Clients (administrators) $1,\dots,K$;
                 Local password reuse graphs $\mathcal{G}_{k} = (\mathcal{V}_{k}, \mathcal{E}_{k}), \forall k \in K$; Training steps $T$; Learning rates $\eta^t, \forall t \in T$}
        \Ensure \rev{Trained GNN weights}
        \rev{\For{step $t \in T$}
            \For{client $k \in K$}
                \Statex \hspace{1cm} \textit{(Client side)}
                \State $\mathbf{w}_{k}^{t+1} = \mathbf{w}_k^{t} - \eta^t \nabla_{\mathbf{w}_k} \mathcal{L}_{k}(\mathbf{w}_{k}; \mathcal{G}_k)$
                \State Transmit $\mathbf{w}_{k}^{t+1}$ to central server $S$.
            \EndFor
            \Statex \hspace{0.5cm} \textit{(Sever side)}
            \State $\mathbf{w}^{t+1} = \frac{1}{K} \sum_{k=1}^{K} \mathbf{w}_{k}^{t+1}$
            \State Transmit the copies of $\mathbf{w}^{t+1}$ to every client.
        \EndFor}
    \end{algorithmic}
    \label{alg:fl}
\end{algorithm}

\jh{
In the inference phase (forward pass), each administrator first computes local node representation vectors using Algorithm~\ref{alg:learning}. These vectors are then shared across administrators, enabling the construction of edge representations between nodes owned by different parties. Specifically, the edge representation $\mathbf{h}_{e_{uv}}$ is derived by concatenating the node embeddings of $u$ and $v$, followed by a linear transformation and activation:
\begin{align}
\mathbf{h}_{e_{uv}} = \sigma(\mathbf{W}_f \cdot \textsc{Concat}(\mathbf{h}_u, \mathbf{h}_v)),
\end{align}
where $\sigma$ is an activation function and $\mathbf{W}_f$ is the weight matrix for edge representation learning. To predict whether an actual password reuse relationship exists between $u$ and $v$, we apply a feed-forward network $f$ followed by a sigmoid activation function:
\begin{align}
\hat{p}_{e_{uv}} = \textit{sigmoid} (f(\mathbf{h}_{e_{uv}})).
\label{eq:probability}
\end{align}
Here, $\hat{p}_{e_{uv}}$ denotes the predicted probability of an edge. A threshold $\tau$ is applied such that a reuse relation is considered to exist if and only if $\hat{p}_{e_{uv}} \ge \tau$. In our implementation, $f$ is a simple fully connected layer. Unlike prior FL-GNN approaches~\cite{scardapane2020distributed,yao2023fedgcn,he2021fedgraphnn}, our method avoids message passing across administrative domains. Instead, cross-admin edges are treated solely as prediction targets during training. This design choice ensures that no password reuse information is directly exchanged across administrators, thereby preserving user privacy.
}

\jh{
In the training phase, our goal is to optimize the global model parameters  
\begin{align}
\mathbf{w} = \{\mathbf{W}^{m \in \{1, ..., M\},\, l \in \{1, ..., L\}},\, \mathbf{W}_f\},
\end{align}
which are initially stored on a central server and distributed to all participating administrators. The optimization objective is defined as:
\begin{align}
    \mathbf{w}^{*} &= \mathop{\arg\min}_{\mathbf{w}_{1}, ..., \mathbf{w}_{K}} \frac{1}{K} \sum_{k = 1}^{K} \mathcal{L}_{k} (\mathbf{w}_{k}; x_{k}; \hat{y}_k, y_k) \nonumber \\ 
    &\text{s.t.} \quad \mathbf{w} = \mathbf{w}_1 = \mathbf{w}_2 = \cdots = \mathbf{w}_K,
\end{align}
where $x_k$ denotes the node features of $\mathcal{V}_k$, and $\hat{y}_k$ and $y_k$ are the predicted and ground-truth labels of password reuse edges $e \in \mathcal{E}_k$ within the local graph. Following the distributed GCN training strategy~\cite{scardapane2020distributed}, this problem can be solved in a fully distributed manner. \rev{Algorithm~\ref{alg:fl} describes the details of the federated learning process.} Each administrator independently minimizes its local loss function $\mathcal{L}_k$ using only its local password reuse graph $\mathcal{G}_k$. The local model update at training step $t$ is computed as:
\begin{align}
    \mathbf{w}_{k}^{t+1} = \mathbf{w}_k^{t} - \eta^t \nabla_{\mathbf{w}_k} \mathcal{L}_{k}(\mathbf{w}_{k}; x_{k}; \hat{y}, y),
\end{align}
where $\eta^t$ is the learning rate at step $t$. After local training, the updates from all administrators are sent to the central server, which aggregates them to update the global model as follows:
\begin{align}
    \mathbf{w}^{t+1} = \frac{1}{K} \sum_{k=1}^{K} \mathbf{w}_{k}^{t+1}
\end{align}
For the local loss function $\mathcal{L}_k$, we use binary cross-entropy between the predicted probability $\hat{p}_{e}$ and the ground truth label $y_{e} \in \{0, 1\}$:
\begin{align}
    \hspace{-0.25cm}
    \mathcal{L}_k = -\frac{1}{|\mathcal{E}_k|} \sum_{e \in \mathcal{E}_k} y_{e} \log \hat{p}_{e} + (1 - y_{e}) \log (1 - \hat{p}_{e}).
\end{align}
}

\noindent\textbf{Mini-batch training and implementation.} We train all weights of the embedding layers (e.g., lookup embedding, LSTM, and batch normalization layers), the GNN layers, and the final feed-forward network in an end-to-end manner. Fully training a large password reuse graph is difficult due to GPU memory requirements (proportional to the number of nodes~\cite{dgldocs}). To handle this problem, we utilize mini-batches of edges in a graph and recursively train every sub-graph which is composed of $1$ to $L$-hop neighborhood of two nodes of a given password reuse relation edge. While edge sampling and sub-graph building tasks are CPU-bound, training each sub-graph is conducted on a GPU. We implement \oursNew{} using Deep Graph Learning (DGL)~\cite{wang2019deep}, a library providing deep learning modules for graphs.

\section{Evaluation}

\jh{In this section, we present our experiment setup on a real-world dataset and the evaluation results of \oursNew{}. We primarily compare \oursNew{} with its original design, \ours{}~\cite{kim2024passrefinder}. Additionally, we provide the extended results of the FL setting, including the effectiveness of tailored GNN models, an ablation study on the modeling choices, and a risk score analysis.}

\subsection{Experiment Setup}

\subsubsection{Dataset}
\label{subsubsec:dataset}

\rev{We adopt the \texttt{Cit0day}~\cite{cit0day1,cit0day2} dataset, which is a large-scale credential dataset from a credential selling service leak in November 2020.} The dataset contains 360,115,257 accounts (i.e., email addresses and passwords) breached from 22,378 websites in the real world. Since features of defunct websites cannot be extracted from public web analytics, we filter them out and leave 19,062 reachable websites. Among them, we first select 15,636 websites where users' plain-text passwords are identifiable to obtain precise information on password reuses.

\begin{figure}
    \centering
    \begin{subfigure}[t]{0.49\linewidth}
        \centering
        \includegraphics[width=\linewidth]{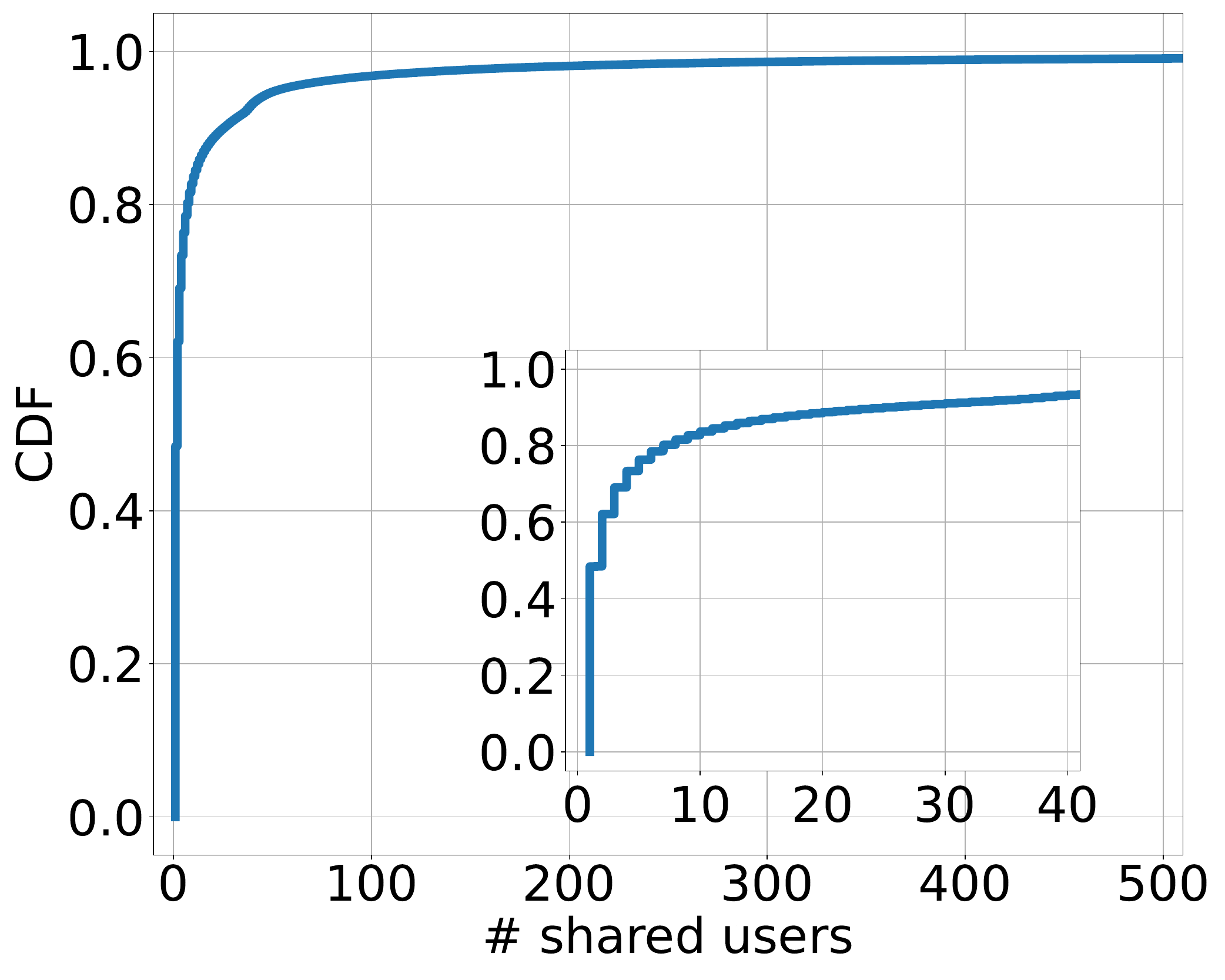}
        \caption{The number of shared users}
        \label{fig:shared_user_distribution}
    \end{subfigure}
    \begin{subfigure}[t]{0.49\linewidth}
        \centering
        \includegraphics[width=\linewidth]{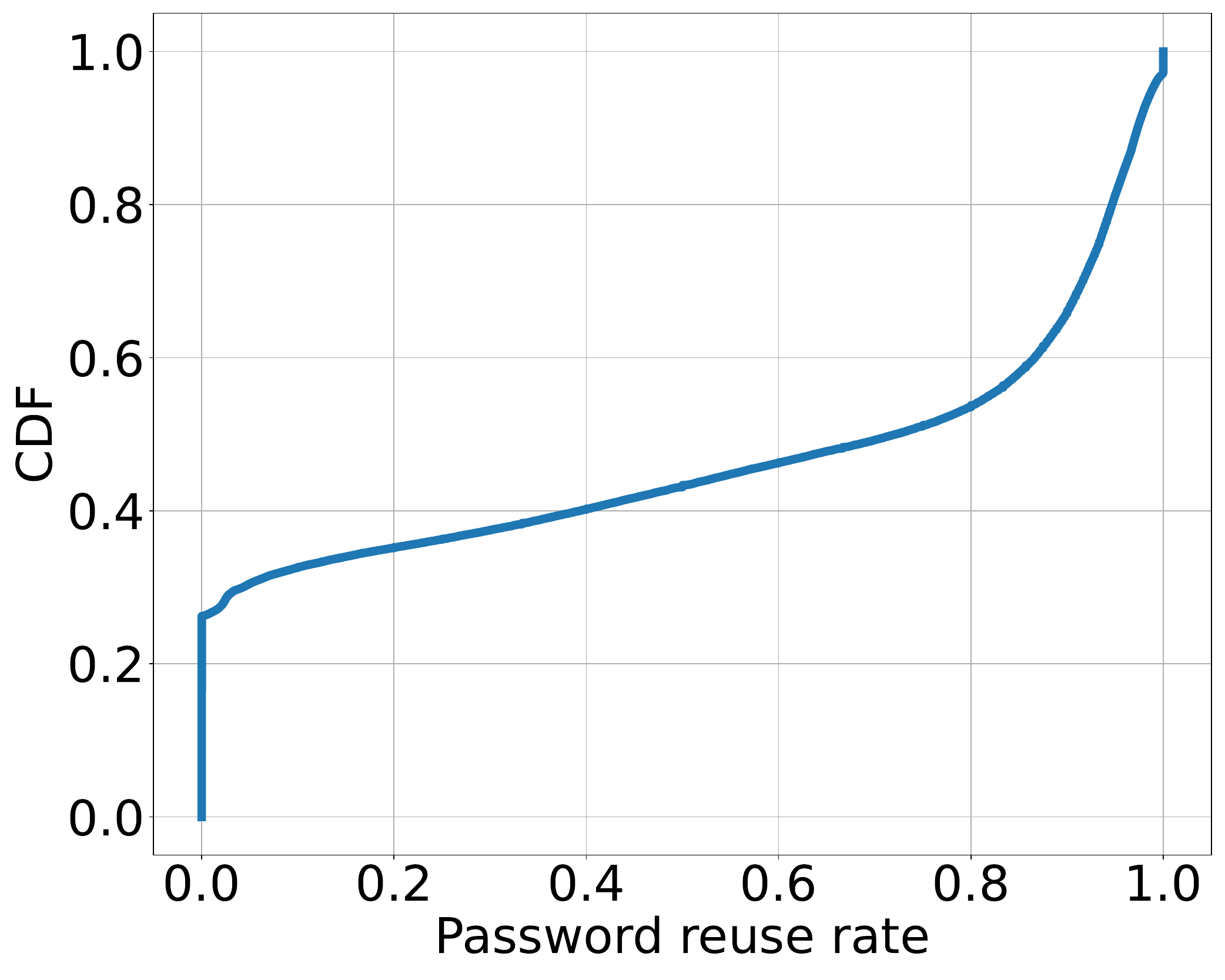}
        \caption{Password reuse rate}
        \label{fig:password_reuse_ditribution}
    \end{subfigure}
    \caption{The distributions of the number of shared users and password reuse rate between websites.}
    \label{fig:distribution}
\end{figure}

A single user is indicated by a unique email address (a commonly used user identifier in password analyses~\cite{das2014tangled, thomas2017data, wang2019end, wang2020detecting}). We note that email addresses are entirely anonymized to address ethical considerations. The number of shared users between websites ranges from 0 to 733,300 with most within 500 (see Figure~\ref{fig:shared_user_distribution}). To ensure the confidence of analysis, we define the ground truth of password reuse relation edges only when the number of shared users is at least 30, which is the minimum number of observations required for statistical significance~\cite{hogg1977probability}. We filter out websites that share fewer than 30 users with every other website, leaving 9,927 websites in the end. Among the shared users between two given websites, the password reuse relation edge is labeled as positive if the rate of users who reuse passwords in the websites is higher than the threshold $\tau$, or labeled as negative otherwise. Website pairs having a zero rate account for 26.2\% of the total (see Figure~\ref{fig:password_reuse_ditribution}). \rev{Consistent with our prior work, \ours{}, we set the threshold $\tau$ to 0.5 in the main experiment.}

\begin{table}
    \centering
    \caption{\jh{Dataset statistics. We present the number of edges in each split set. \textit{Local} edges refer to those within each local password reuse graph $\mathcal{G}_k = (\mathcal{V}_k, \mathcal{E}_k), k \in \{1, ..., K=10\}$. \textit{Cross-admin} edges represent those connected across different administrators' websites and serve as the prediction targets in our experiment.}}
    \footnotesize
        \begin{tabular} {l l r}
            \toprule
            \textbf{Edge type} & \textbf{Split} & \textbf{\# of edges} \\ \midrule
            Local ($\mathcal{E}_k$) & Train & \makecell{48,990/14,484/16,000/15,144/11,242\\/7,594/14,110/23,466/5,842/9,820} \\ \midrule
            \multirow{2}{*}{Cross-admin} & Valid  & 24,421 \\
            & Test        & 498,204 \\ \midrule
            Total           & & 605,971 \\
            \bottomrule
        \end{tabular}%
    \label{tab:dataset}
\end{table}

\jh{Our final dataset for the experiments consists of 9,927 website nodes connected by 605,971 password reuse relation edges. To simulate an FL setting with $K = 10$ participating administrators, we partition the original graph into ten disjoint subgraphs. Each subgraph corresponds to one administrator, with 1,000 nodes assigned per administrator, except for the last subgraph containing 927 nodes. As a result, all cross-admin password reuse relation edges are removed during partitioning and excluded for message passing in the GNN model's forward pass. For evaluation, we use cross-admin edges between two randomly selected administrators as the validation set, while the remaining cross-admin edges are used for testing. Detailed statistics of the dataset are presented in Table~\ref{tab:dataset}.}

\subsubsection{Training and Hyperparameters}

In our prototype framework, we set the dimension $D_{m}$ of the lookup embedding vectors and the LSTM output vectors to 256. We use the embedding vector of XLM-RoBERTa with the default 768 dimensions. For each of the $M = 5$ modalities, we train $L = 2$ layers of the GNNs with $d = 256$ dimensions of hidden vectors. In each GNN layer, we leverage all neighbors of each node, instead of sampling from them, to fully learn the neighborhood information. We also adopt leaky ReLU with 0.2 negative slope for the activation function $\sigma$ and a fully connected layer for the feed-forward network $f$ as our design choice.

We evaluate various settings of hyperparameters: maximum epochs of 100, 200, 500, and 1,000; (edge) batch sizes of $2^{12}$, $2^{14}$, $2^{16}$, and $2^{18}$; maximum learning rates of $10^{-4}$, $10^{-3}$, $10^{-2}$, and  $10^{-1}$. Then, we select a maximum epoch of 200 with 40 epochs for early stopping, a batch size of $2^{16}$, and a maximum learning rate of $10^{-3}$ using the Adam optimizer, of which combination shows the best performance. In addition, 1 cycle policy~\cite{smith2019super} with a warmup of 10\% is adopted to schedule the learning rate. We evaluate our framework on a machine with 20 Intel i9-10900K CPU cores, 128GB of memory, and a single NVIDIA RTX 3090 GPU.

\begin{table*}
    \centering
    \caption{\jh{Overall prediction performance. The best and the second best scores are in bold and underlined, respectively.}}
    \footnotesize
        \begin{tabular} {l c c c c c c}
            \toprule
            \multirow{2}{*}{\textbf{Model}} & \multirow{2}{*}{\textbf{Hidden relation policy}} & \multirow{2}{*}{\textbf{Privacy-preserving}} & \multicolumn{3}{c}{\textbf{Performance}} \\ \cmidrule{4-6}
            & & & \textbf{Precision} & \textbf{Recall} & \textbf{F1-score} \\ \midrule
            \multirow{3}{*}{\ours{}~\cite{kim2024passrefinder}} & \textit{No hidden relation} & \checkmark & 0.9238 & 0.6405 & 0.7565  \\
            & Website feature similarity ($Sim_{avg}$) & \checkmark & 0.8851 & 0.6971 & 0.7799\\
            & Overlapped user number ($User_{30}$) & \xmark & \textbf{0.9484} & \underline{0.7274} & \underline{0.8234} \\ \midrule
            \oursNew{} & \textit{No hidden relation} & \checkmark & \underline{0.9425} & \textbf{0.8896} & \textbf{0.9153} \\
            \bottomrule
        \end{tabular}%
    \label{tab:overall}
\end{table*}

\subsubsection{Baselines}

\noindent\jh{
\textbf{\ours{}~\cite{kim2024passrefinder}.} We evaluate the credential stuffing risk prediction framework proposed in our prior work, which leverages hidden relation modeling to estimate risks across websites managed by different administrators. Specifically, we assess three variants based on different hidden relation policies: \textit{No hidden relation}, $\ours{}_{Sim_{avg}}$, and $\ours{}_{User{30}}$\footnote{The $\ours{}_{Sim_{avg}}$ model creates a hidden relation edge if the feature similarity between two websites exceeds the average similarity observed in the training set. The $\ours{}_{User{30}}$ model creates an edge when at least 30 shared users exist, providing statistical significance~\cite{kim2024passrefinder}.}. For training \ours{}, we use only the first local password reuse graph $\mathcal{G}_1$, which consists of 1,000 nodes and 48,990 edges (see Table~\ref{tab:dataset}).
}

\noindent\jh{\textbf{GNN variants with FL.} Under the same FL setting, we compare various GNN model architectures, which have been frequently used in risk prediction tasks~\cite{bilge2017riskteller, soska2014automatically, veeramachaneni2016ai, he2022illuminati, zhang2020multi}:
\begin{itemize}[leftmargin=*]
    \item \textbf{Graph convolutional network (GCN)} introduced by Kipf et al.~\cite{kipf2016semi} is a GNN with a layer-wise propagation rule and neighbor aggregation based on spectral graph convolutions. We adopt $L = 2$ layers of GCNs for learning representations from multiple hops of neighbors as in our GNN model.
    \item \textbf{Graph attention network (GAT)}~\cite{velivckovic2017graph} is a GCN variant adopting self-attentional neighbor aggregation for selectively learning the influences of important neighbors.
    \item \textbf{GraphSAGE}~\cite{hamilton2017inductive} is a GCN variant for effectively performing inductive learning by concatenating the self-embedding with the neighbor aggregation vector. We set a mean pooling function for neighbor aggregation.
\end{itemize}
\noindent{}For a fair comparison with \oursNew{}, we utilize multi-modality node features rather than relying on the conventional feature concatenation approach.}

\subsection{Results}

\subsubsection{Prediction Performance}
\label{subsec:performance}

\jh{The performance of \oursNew{} and its comparison with the original framework, \ours{}~\cite{kim2024passrefinder}, are summarized in Table~\ref{tab:overall}. \oursNew{} achieves superior prediction performance with an F1-score of 0.9153 while preserving privacy under the FL-based setting. This notable effectiveness is attributed to its ability to comprehensively leverage ground truth (i.e., password reuse information) from websites managed by multiple administrators during both training and inference. In contrast, \ours{} is limited in its access to such information beyond a single administrator’s websites, resulting in lower performance. Among the \ours{} variants, the implementation based on overlapping user information achieves the highest F1-score of 0.8234, consistent with prior results~\cite{kim2024passrefinder}. Notably, this approach requires sharing usernames across potentially untrustworthy administrators, posing significant privacy risks to users. Privacy-preserving implementations maintain better privacy: the one with no hidden relation and the website feature similarity-based hidden relation method. However, their prediction performance is limited due to the smaller size of the available training data.}

\jh{To evaluate the effectiveness of our GNN design within the FL-based framework, we compare several GNN variants under the same setting as \oursNew{}. The experimental results are presented in Table~\ref{tab:variant}. The GNN model used in \oursNew{} achieves the highest F1-score among the variants, effectively balancing precision and recall. Similar to previous observations for \ours{}, GCN and GAT suffer from low precision scores. In contrast, GraphSAGE---known for its robustness in inductive learning settings---outperforms other variants with a competitive F1-score of 0.8790.}

\begin{table}
    \centering
    \caption{\jh{Prediction performance across GNN variants under the FL setting. The best and the second best scores are in bold and underlined, respectively.}}
    \footnotesize
        \begin{tabular} {l c c c c}
            \toprule
            \multirow{2}{*}{\textbf{Model}} & \multicolumn{3}{c}{\textbf{Performance}} \\ \cmidrule{2-4}
            & \textbf{Precision} & \textbf{Recall} & \textbf{F1-score} \\ \midrule
            GCN~\cite{kipf2016semi} & 0.8146 & 0.8243 & 0.8195 \\
            GAT~\cite{velivckovic2017graph} & 0.7732 & \textbf{0.9555} & 0.8547 \\
            GraphSAGE~\cite{hamilton2017inductive} & \textbf{0.9462} & 0.8207 & 0.8790  \\ \midrule
            \oursNew{} & \underline{0.9425} & \underline{0.8896} & \textbf{0.9153} \\
            \bottomrule
        \end{tabular}%
    \label{tab:variant}
\end{table}

\noindent\textbf{Comparison with existing detection methods.} We conduct a comparative analysis between our website-level prediction model and existing user-level detection methods. The detection accuracy of a C3 service can be approximated based on its database coverage. We evaluate the HIBP database before the leak~\cite{hibp} and determine that it detects reused passwords with an accuracy of 0.6510. In the coordination method~\cite{wang2019end}, we evaluate the probability of a user's password being reused in at least one of 64 queried websites: $1 - (\mathbb{P}(\pi = \Psi_i(a)))^{64}$ for password $\pi$ of user $a$, where $\Psi_i(a)$ denotes the password of user $a$ in website $i$. This exhibits an average accuracy of 0.7857 in our dataset. \jh{In addition to the advantages of scalability and minimal privacy implications, \oursNew{} (with 0.9153 F1-score) significantly outperforms these two detection methods.}

\subsubsection{Ablation Study}
\label{subsec:ablation_study}

\jh{
To evaluate the effectiveness of each design strategy, we conduct an ablation study on the GNN modeling components of \oursNew{}, as presented in Table~\ref{tab:ablation}. Specifically, we examine the contributions of the multi-modality feature and the neighborhood attention methods (Section~\ref{sec:design}).
}

\jh{
For the ablation of the multi-modality features, we replace the separate processing of different feature modalities with a simple concatenation of all website node features, followed by a single unified GNN. The results show a substantial drop in prediction performance from 0.9153 to 0.7383, highlighting the importance of multi-modality feature modeling even under the FL setting. This finding is consistent with prior observations in \ours{}~\cite{kim2024passrefinder}.
}

\begin{table}
    \centering
    \caption{\jh{Ablation study on the GNN modeling methods}.}
    \footnotesize
    \begin{tabular} {l c c c}
        \toprule
        \multirow{2}{*}{\textbf{Ablation}} & \multicolumn{3}{c}{\textbf{Performance}} \\ \cmidrule{2-4}
        & \textbf{Precision} & \textbf{Recall} & \textbf{F1-score} \\ \midrule
        Multi-modality & 0.8730 & 0.6396 & 0.7383 \\
        Neighborhood attention & 0.9429 & 0.8606 & 0.8999 \\ \midrule
        \textit{None} (\oursNew{}) & 0.9425 & 0.8896 & 0.9153 \\
        \bottomrule
    \end{tabular}%
    \label{tab:ablation}
\end{table}

\jh{
To assess the impact of the neighborhood attention mechanism, we replace it with a standard mean pooling aggregation. While this method showed clear benefits in the inductive setting of \ours{}, its contribution in \oursNew{} is less significant, with only a slight gain in recall. We attribute this difference to the FL-based approach’s ability to leverage comprehensive website information across multiple administrators, allowing \oursNew{} to maintain remarkable prediction performance without requiring fine-grained prioritization of neighborhood nodes.
}

\subsubsection{\rev{Size of Federation}}

\rev{In federated learning, the number of participating local clients is crucial for achieving satisfactory performance. To evaluate the robustness of \oursNew{} in the federated setup, we examine prediction performance as the number of participating administrators varies.}

\begin{figure}
    \centering
    \includegraphics[width=0.8\linewidth]{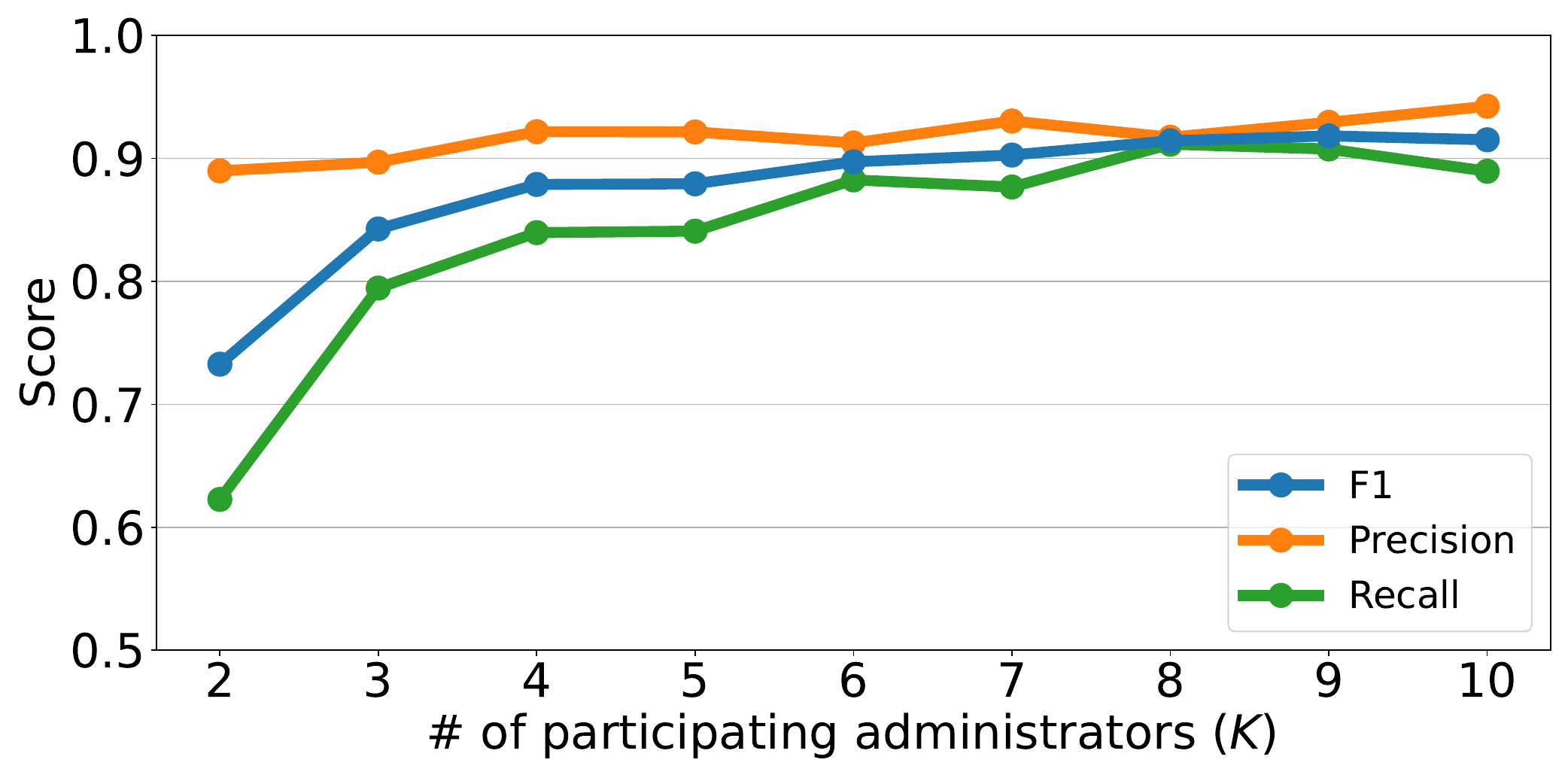}
    \caption{\rev{Prediction performance with varying numbers of administrators participating in federated learning.}}
    \label{fig:nclients}
\end{figure}

\rev{The results are shown in Figure~\ref{fig:nclients}. For fair evaluation, we fix the test set to the cross-admin edges across all local password reuse graphs. Prediction performance, measured by F1, precision, and recall, improves with greater participation. This improvement is attributed to expanded learned knowledge and generalization capability, particularly on cross-admin edges across local password reuse graphs. With only two administrators, the F1 score is 0.7327, falling short of \ours{} (our prior work). Nevertheless, practical performance (F1-scores of >0.9) is achieved with at least five administrators. Consequently, \oursNew{} remains effective with a reasonable federation size, demonstrating practical advantages under limited administrator participation in real-world scenarios.}

\begin{figure}
    \centering
    \includegraphics[width=0.8\linewidth]{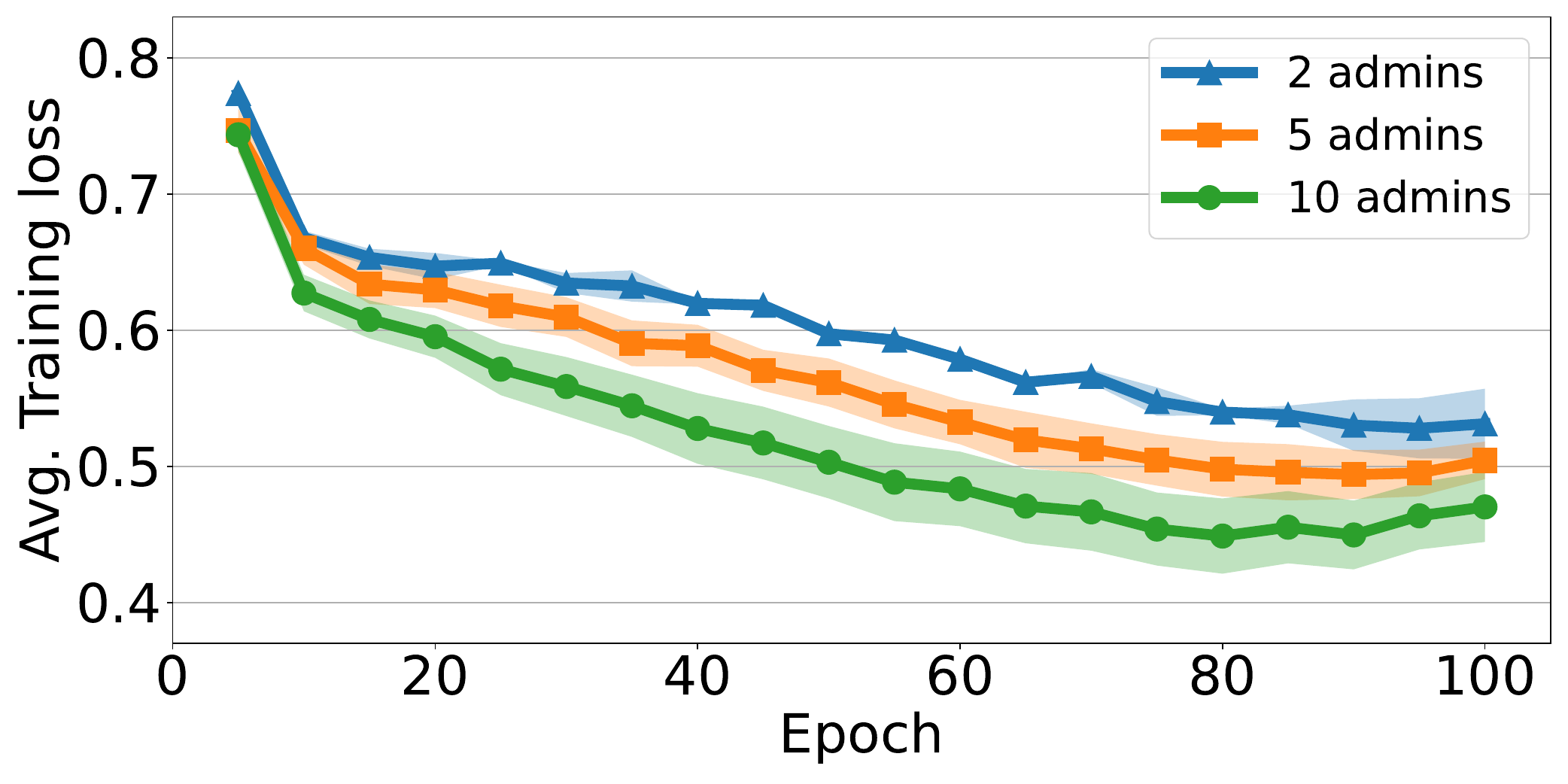}
    \caption{\rev{Training dynamics across federation sizes. The shaded regions indicate standard errors across participating administrators.}}
    \label{fig:training_loss}
\end{figure}

\rev{To further examine the impact of federation size, we track the averaged loss (i.e., $\mathbb{E}_{k \in \{1,\dots,K\}} \mathcal{L}_k$) over training, as shown in Figure~\ref{fig:training_loss}. Overall, \oursNew{} exhibits successful convergence in the federated learning setup, indicating that it learns effective representations of cross-admin password reuse relations from local password reuse graphs. As the number of participating administrators increases, the loss converges more stably and to a lower value, consistent with the improvements in prediction performance in Figure~\ref{fig:nclients}.}

\subsubsection{Risk Score Analysis}
\label{subsubsec:risk_score_analysis}

Some administrators may prefer a more fine-grained outcome for quantifying risk rather than a binary result indicating whether a password reuse rate is high or low. Therefore, we show that the prediction probability $\hat{p}_{e_{uv}}$ in Equation~(\ref{eq:probability}) can serve as \textit{risk scores} of credential stuffing---the expected password reuse rates between two website nodes $u$ and $v$. Specifically, we calculate the average risk scores between a given website and the others to obtain the website's risk score. We examine two tasks: prioritizing website pairs by their relative risks and estimating the risk score value. For comparison, we present the results of \ours{} with the user information-based hidden relation method ($User_{30}$).

\begin{figure}
    \centering
    \begin{subfigure}[t]{0.45\linewidth}
        \centering
        \captionsetup{justification=centering}
        \includegraphics[width=\linewidth]{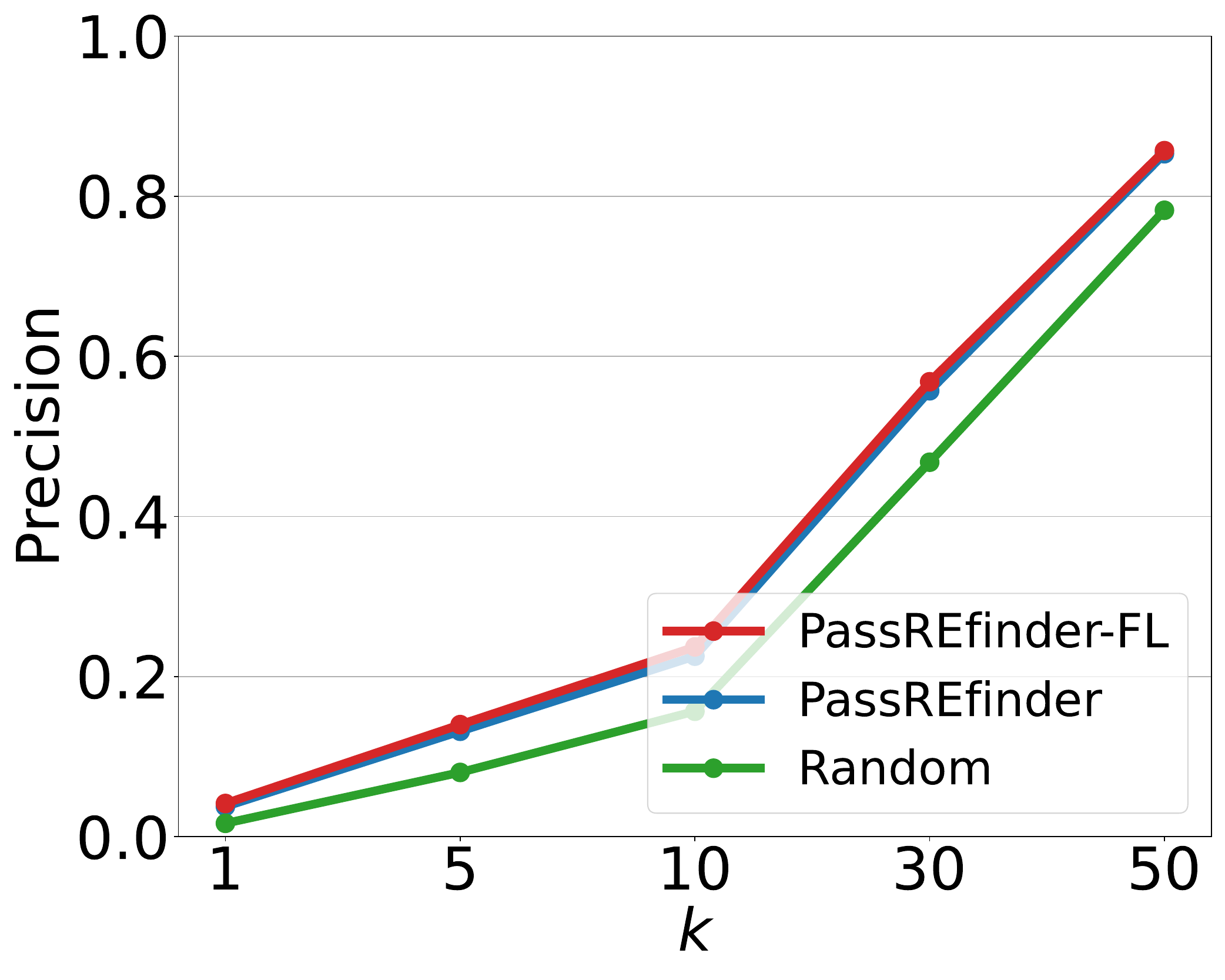}
        \caption{Precision@$k$}
        \label{fig:ranking_prec}
    \end{subfigure}
    \hspace{1mm}
    \begin{subfigure}[t]{0.45\linewidth}
        \centering
    \captionsetup{justification=centering}
        \includegraphics[width=\linewidth]{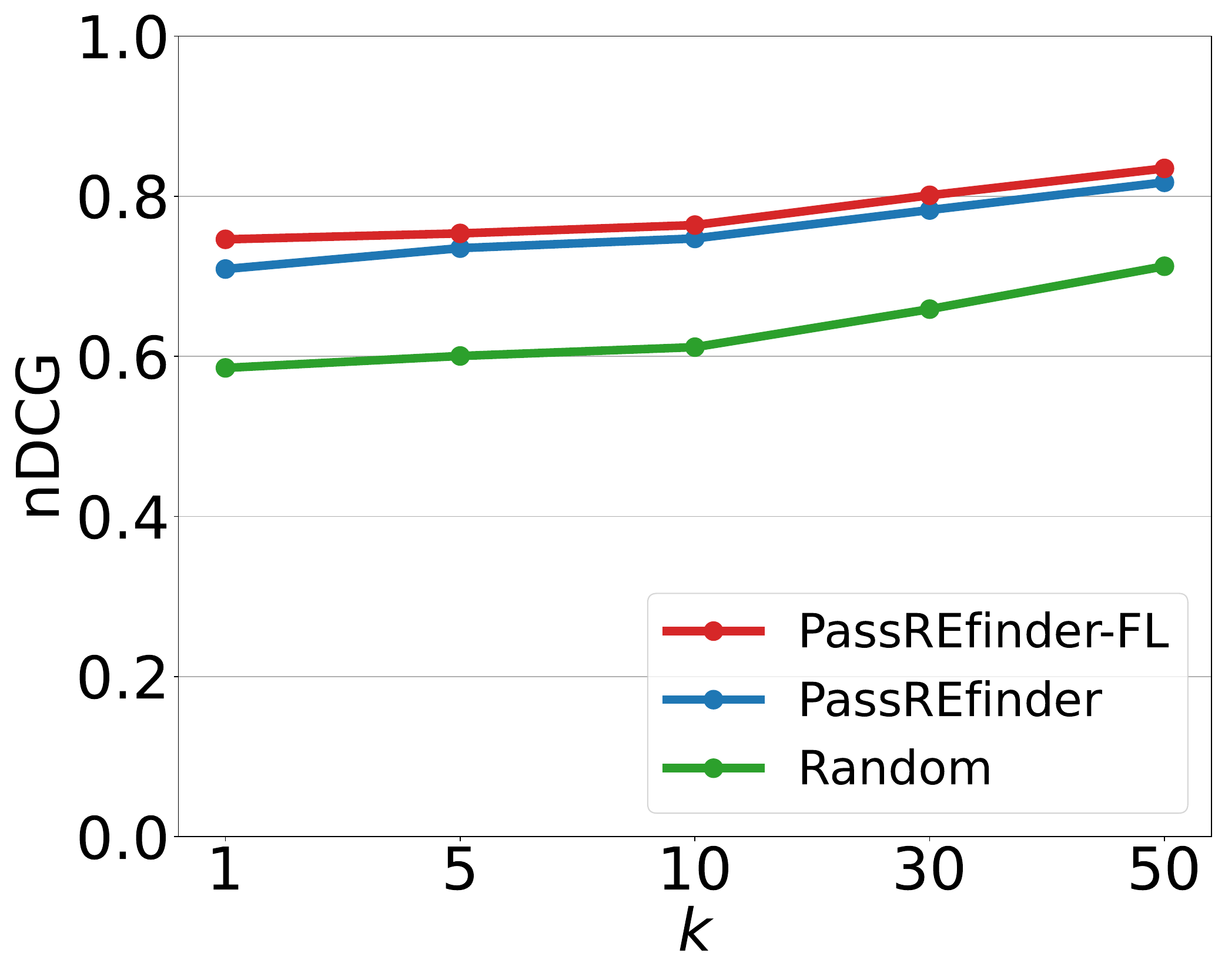}
        \caption{nDCG@$k$}
        \label{fig:ranking_ndcg}
    \end{subfigure}
    \caption{\jh{Evaluation using ranking metrics. We present the results with 64 sampled edges for each website node.}}
    \label{fig:ranking}
\end{figure}

\jh{
We first demonstrate that the prediction probability produced by our prediction model effectively prioritizes the risk levels between pairs of websites. To evaluate this, we use ranking-based metrics: precision and normalized discounted cumulative gain (nDCG) at top-$k$ predicted edges for a given website node (see Table~\ref{fig:ranking}). Precision@$k$ measures the proportion of truly positive edges among the top-$k$ predicted edges, while nDCG@$k$ considers both the relevance (i.e., risk score) and the ranking order. As a baseline, we introduce a naive method (\textit{Random}) that randomly predicts the risk score between two websites based on a uniform distribution.
}

\jh{
\oursNew{} significantly outperforms the \textit{Random} baseline in both precision@$k$ and nDCG@$k$, demonstrating its effectiveness in prioritizing website pair across different administrators by identifying higher-risk password reuse relations. \ours{} performs much worse than \oursNew{} in the link prediction task (as shown in Table~\ref{tab:overall}). In contrast, it serves as a competitive baseline for risk score prediction, where \oursNew{} achieves only marginal improvements. Nonetheless, the primary advantage of \oursNew{} lies in its privacy-preserving design, making it highly practical for real-world deployment without requiring local password reuse information sharing between administrators.
}

\begin{figure}
    \centering
    \begin{subfigure}[t]{0.45\linewidth}
        \centering
        \captionsetup{justification=centering}
        \includegraphics[width=\linewidth]{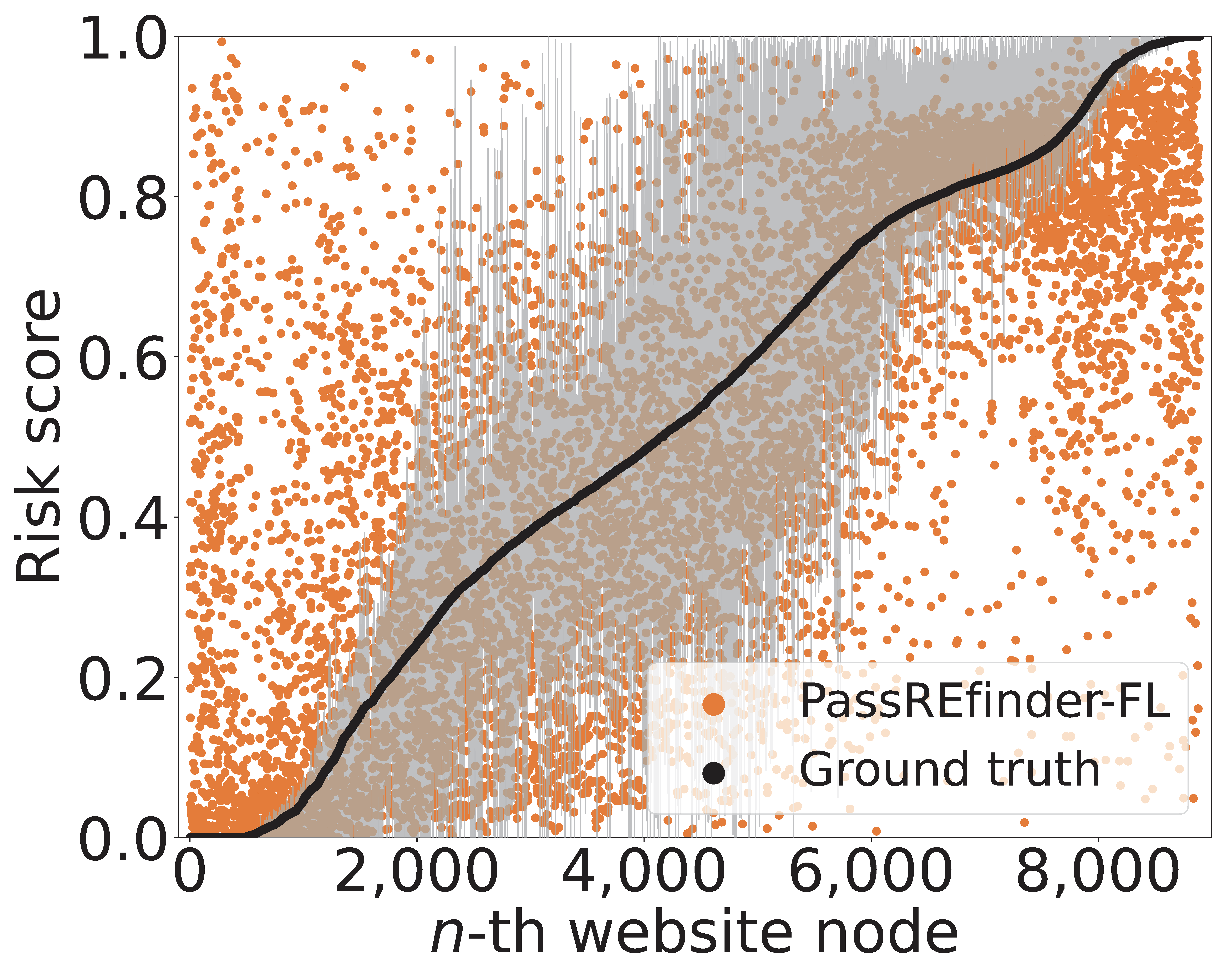}
        \caption{\oursNew{}\\$MSE$=0.0581}
        \label{fig:risk_score_fl}
    \end{subfigure}
    \hspace{1mm}
    \begin{subfigure}[t]{0.45\linewidth}
        \centering
    \captionsetup{justification=centering}
        \includegraphics[width=\linewidth]{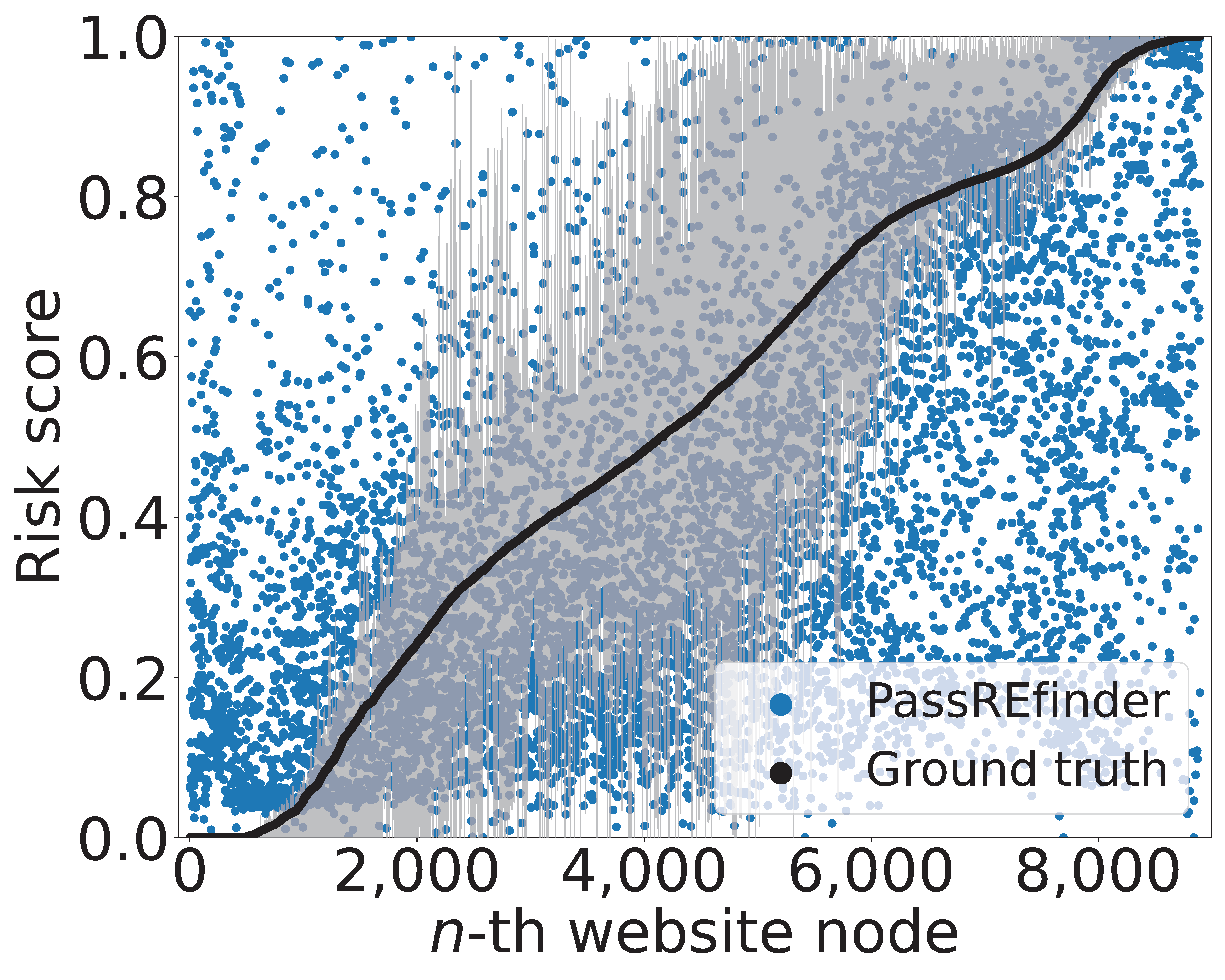}
        \caption{\ours{}\\$MSE$=0.0925}
        \label{fig:risk_score_old}
    \end{subfigure}
    \caption{\jh{Risk scores averaged over the edges from each website node. The shaded regions are the range between the 25th and 75th percentiles of the ground truth risk scores. $MSE$ denotes the mean squared error of each result.}}
    \label{fig:risk_score}
\end{figure}

\jh{
Additionally, we examine whether the prediction probability can be directly interpreted as a risk score. In Figure~\ref{fig:risk_score}, we visualize the prediction probabilities generated by \ours{} and \oursNew{}, along with their errors from the true password reuse rates (i.e., Ground truth). \oursNew{}’s predictions closely align with the ground truth, achieving slight improvements over \ours{}. These results suggest that our GNN model, under the FL setting, can quantitatively measure credential stuffing risks between websites in practical scenarios.
}

\subsubsection{\rev{Communication Cost Analysis}}

\rev{
We theoretically estimate the communication overheads of \oursNew{} in the training and inference processes, respectively.
}

\noindent\rev{\textbf{Training.} Every federated learning round $t$, the participating administrators $k \in \{1,\dots,K\}$ upload their local updates and download the aggregated global parameters $\mathbf{w}$ (see Algorithm~\ref{alg:fl}). 
Here, the total model size is computed as 
$|\mathbf{w}|=\sum_{m,l}|\mathbf{W}^{m,l}|+|\mathbf{W}_f|$.
The total communication cost of training is:
\begin{align}
C_{\text{train}} &= \sum_{t=1}^{T} \Big( C_{\text{up}}^{(t)} + C_{\text{down}}^{(t)} \Big) 
\approx 2\,b\,|\mathbf{w}| K T,
\end{align}
where $b$ denotes the number of bytes per scalar.
which grows linearly with the number of participating administrators and rounds. Since $|\mathbf{w}|$ is fixed, the overhead remains predictable and scalable, ensuring practicality even in large federations.
}

\noindent\rev{\textbf{Inference.} At the inference time, cross-admin prediction requires only transmitting node embedding vectors. For a test edge set $\mathcal{Q}$ with embedding dimension $d$, the communication cost is:
\begin{align}
C_{\text{infer}}(\mathcal{Q}) = \delta\, b\, d\, |\mathcal{Q}|,
\end{align}
where $\delta\in{1,2}$ indicates whether the implementation transmits embedding vectors in one way ($\delta=1$) or bidirectionally ($\delta=2$). In line with the communication cost during training, this cost scales linearly with the size of the test dataset and the embedding dimension, and is significantly lighter than exchanging raw graphs.
}

\rev{
Overall, \oursNew{} achieves communication efficiency by design: both training and inference overheads increase only \textit{linearly} with federation sizes or test dataset sizes, highlighting its scalability and practicality for real-world deployment.
}
\section{\rev{Practical Implementation and Application of \oursNew{}}}
\label{sec:application}

\begin{figure}
    \centering
    \includegraphics[width=\linewidth]{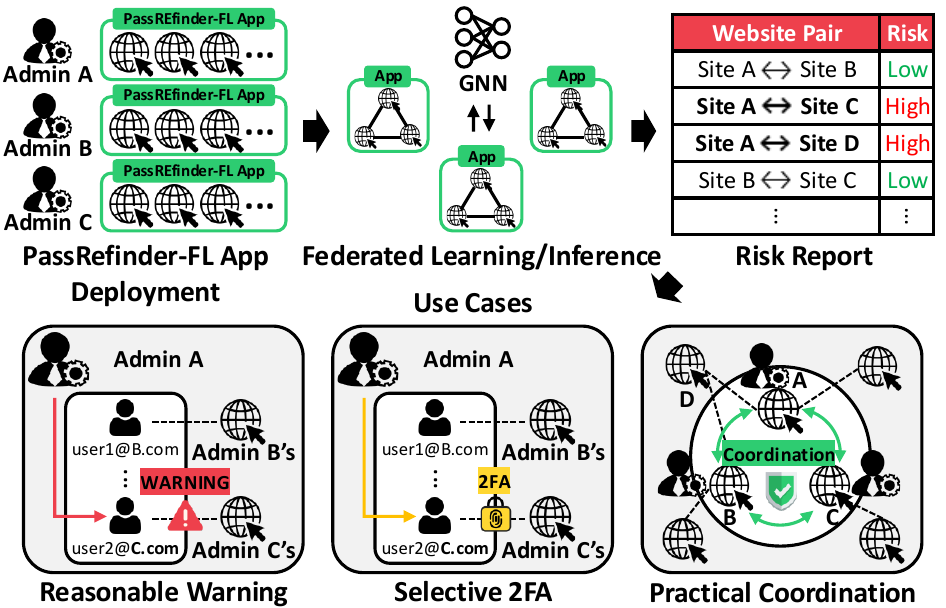}
    \caption{\rev{\oursNew{} workflow and its practical use cases.}}
    \label{fig:workflow}
\end{figure}

\rev{We illustrate the workflow of \oursNew{} in Figure~\ref{fig:workflow}. \oursNew{} can be implemented as an application (i.e., \oursNew{} App), and administrators deploy the \oursNew{} App in their local environments by linking the websites’ credential databases. For a secure and privacy-preserving implementation, administrators should pseudonymize usernames and hash passwords before providing them to the \oursNew{} App. In each administrator’s local environment: i) the \oursNew{} App explores the credential databases of the administrator’s websites and assesses credential stuffing risks among them. ii) It constructs a local password reuse graph and extracts website features from public web analytics services. iii) Under federated learning, it trains the local GNN model using local credential stuffing risk information in a privacy-preserving manner. iv) By transmitting only the inferred embedding vectors, it predicts credential stuffing risks between websites, particularly across different administrators (i.e., cross-admin password reuse relations). This process produces a credential stuffing risk report that specifies the risk level for each website pair (i.e., whether the risk is high or low). Based on these results, administrators can identify \textit{risky websites} that pose potential threats to their local websites. Notably, this preemptive risk prediction approach addresses the limited detection coverage of C3 services and the poor password usability and scalability of coordination methods. Consequently, administrators can proactively mitigate credential stuffing risks using insights into risky websites. We describe three practical use cases of \oursNew{}:}

\noindent\textbf{Reasonable warning.} \rev{One use case of \oursNew{} is to selectively warn users of high-risk relationships between administrators' websites and risky websites. To illustrate this, consider the case where \oursNew{} is not deployed but administrators are aware of risky websites (e.g., Admin C's website in Figure~\ref{fig:workflow}). A conventional approach is for administrators to broadcast password-reset warnings to \textit{all users} across their managed websites. However, this blanket strategy relies heavily on users' security awareness and may lead to fatigue from repeated exposure to warnings. Another existing approach is password reuse notifications (e.g., Google Password Manager~\cite{thomas2019protecting}), which inform users when the same password is reused from breached websites. Yet, such tools do not estimate the risk of password reuse relations and are not proactive solutions.}

\rev{
\oursNew{} instead provides \textit{selective warnings}, targeting only specific users based on estimated risks. It first infers potentially affected users by matching email domains with risky website domains, thereby avoiding the impracticality of directly identifying user registration data. Then, \oursNew{} delivers warnings exclusively to those at risk, accompanied by detailed reasoning evidence (e.g., website features). This strategy not only reduces the overhead of broadcast warnings but also enhances users’ security awareness through targeted education.}

\noindent\textbf{Selective 2FA.} \rev{Two-factor authentication (2FA) has been widely deployed to defend against credential-related attacks. However, enforcing 2FA for all users on a website is often impractical due to the required effort and time~\cite{colnago2018s, golla2021driving}. Instead, selectively activating 2FA for specific users based on our prediction results can strike a balance between user costs and credential stuffing mitigation. To this end, \oursNew{} enables administrators to adopt an approach similar to the selective warning use case. \oursNew{} first infers users with accounts on risky websites by matching their email domains with those websites. It then selectively recommends or enforces 2FA adoption for these users, providing detailed explanations of website features and associated risks to encourage compliance. Furthermore, administrators can apply fine-grained levels of 2FA enforcement based on the risk scores evaluated by \oursNew{}, thereby allowing careful consideration of the trade-off between user convenience and strong protection.}

\noindent\textbf{Practical coordination.} Although website coordination sharing usernames and passwords is a promising approach to combat credential stuffing, existing implementations of website coordination~\cite{wang2019end, wang2020detecting} have suffered from scalability and reluctance from administrators to participate. \oursNew{} seeks to overcome these challenges by creating modestly sized yet effective coordination pools. To this end, administrators i) share prediction results to select each website set effective for coordination, where each website within a set has credential stuffing risks with the others. They can ii) provide persuasive explanations, underpinned by website features, to encourage other administrators' active participation in the coordination process. Then, administrators iii) deploy coordination protocols for detecting password reuse or credential stuffing per each website set. While addressing the scalability problem, \oursNew{} can significantly improve the detection accuracy of coordination methods, as indicated by our comparison results in Section~\ref{subsec:performance}.

\section{Limitation and Discussion}

\noindent\rev{\textbf{Reliability of website features.}} The feature extraction process yields valuable website features for modeling password reuse tendencies. Nonetheless, discrepancies may arise between a user's password creation time and the feature extraction time, potentially affecting the ground truth password reuse rate. To minimize this disparity, even in the absence of precise dates of website creation or user registration, we focus on active websites and randomly split the dataset for our training and testing processes.

In addition, multiple domains within a single cloud hosting server could exhibit varying security postures. In the absence of administrative privileges, our framework relies on server-specific security postures to achieve the capability of deployment on a vast number of arbitrary websites. It is worth noting that the effectiveness of server-specific security postures for the risk prediction task was demonstrated through our ablation studies. Nevertheless, we could further enhance the reliability of our framework by achieving domain-specific security postures. In addition, a single cloud-hosting server could host multiple domains, each exhibiting unique security postures. In scenarios where administrative privileges are not accessible, our framework adopts the server-specific security postures of websites. This enables the framework's deployment across an extensive range of arbitrary websites.

It is noteworthy that the evaluation results demonstrate the effectiveness of server-specific security postures in predicting credential stuffing risk. Still, the reliability of our framework could be further enhanced by achieving domain-specific security postures. The IP addresses of cloud hosting servers also may not accurately represent website service locations. Employing \textit{Shodan}, we ascertain that only 13.6\% of websites in our dataset are hosted by major cloud providers, and conservatively estimate that the hosting countries of one-third of these websites correspond with their website service countries, as inferred from their top-level domains. Furthermore, the multi-modal attention can penalize insignificant features (e.g., global website locations). Consequently, we posit that these limitations exert minimal impact on our findings.

\noindent\rev{\textbf{Effectiveness in asymmetric federation.} Our experiments assume a balanced federation setup in which each local password reuse graph contains 1,000 nodes. This setting may not fully reflect real-world scenarios where the number of websites associated with an administrator varies significantly. In this asymmetric federation setup, the prediction capability of \oursNew{} can be generalized to major administrators’ websites by benefiting from sufficient information propagation across their large local password reuse graphs, while performance may degrade for minor administrators with small local password reuse graphs. Nevertheless, we expect the practical impact of this limitation on risk prediction to be negligible because credential stuffing risks associated with major administrators’ websites are more crucial and broadly influential in practice and should be prioritized.}

\noindent\rev{\textbf{Risks of credential tweaking.} Several studies have examined a related vulnerability in which users slightly modify their passwords instead of reusing them verbatim, known as credential tweaking~\cite{das2014tangled,pearman2017let,wang2018next,pal2019beyond,xu2023improving,wang2023pass2edit}. Nevertheless, our default design focuses on credential stuffing for two reasons: i) Credential stuffing is more practical and has broader impact in real-world settings. Credential tweaking typically requires multiple guesses and repeated login attempts guided by a modification rule or a password guessing model, which imposes substantial overhead and limits large scale untargeted attacks. For example, a simulation of credential tweaking~\cite{pal2019beyond} shows that gaining an additional $+15.8\%p$ in success rate requires about $1{,}000\times$ more login attempts with the prediction cost of the guessing model; attackers may prefer credential stuffing, which can achieve about a $40\%$ success rate with a single login attempt per user. ii) The risk of credential stuffing can be quantified systematically, enabling concrete evaluation in practice. In contrast, credential tweaking depends on chosen modification rules or guessing models, making a consistent and robust definition of \textit{reuse} difficult. Therefore, we estimate a conservative lower bound on risk from password reuse by restricting our scope to credential stuffing. Nonetheless, \oursNew{} can be extended to predict the risks of credential tweaking by explicitly defining \textit{password reuse} through a selected password modification rule or a password guessing model.}

\section{Related Work}
\label{sec:related_work}

\noindent\textbf{Compromised credential checking (C3) services.} C3 services allow users to cross-reference usernames and passwords based on leaked credential datasets~\cite{hibp, thomas2019protecting, microsoft2021}. They also help users deal with credential stuffing attacks using breach-alerts for their passwords. Thomas et al.~\cite{thomas2019protecting} presented Google Password Checkup (GPC), a protocol for querying usernames and passwords from credential data breaches. Li et al.~\cite{li2019protocols} formally described C3 services and their security requirements. Pal et al.~\cite{pal2022might} extended a C3 service to target credential tweaking attacks~\cite{pal2019beyond}. However, existing C3 services cannot prevent attacks that exploit new accounts from undisclosed credential data breaches and suffer from adversarial exploits to extract credential data~\cite{pal2022might}.

In contrast, our proposed framework predicts the potential risk of credential stuffing residing between arbitrary websites. Therefore, we can prevent credential stuffing attacks before accounts are breached without relying on unreliable databases of leaked accounts.

\noindent\textbf{Coordination for detecting credential stuffing.} Website coordination has emerged to actively protect user accounts~\cite{wang2019end, wang2020detecting}. As a pioneering work, Wang et al.~\cite{wang2019end} proposed a framework for preventing a user in a \textit{requester} website from setting the same password in \textit{responder} websites. They developed an improved private set-membership-test (PMT) protocol underlying their framework to minimize the leakage of user information. Unfortunately, the coordination method degrades the usability of passwords by interrupting password choice. Wang et al.~\cite{wang2020detecting} mitigated this problem by modifying the design to directly detect credential stuffing attacks. Their protocol relied on a black-box anomaly detection system such as traffic fingerprinting techniques~\cite{herley2019distinguishing, tian2020stopguessing, seo2022heimdallr}. However, the cost of false positives in detection is expensive due to the immediate denial of user access to websites. While they claimed that their protocols were able to handle 64 or 256 \textit{responder} websites, scaling to millions or billions of users was infeasible~\cite{pal2022might}. Moreover, their username-website mapping systems are problematic in negotiating the coordination of unreliable websites.

Our proposed framework predicts the potential risk of credential stuffing attacks between websites before any disruptions can occur to users, and enables the deployment of coordination protocols by selecting effective \textit{responder} websites for given a \textit{requester} website. Furthermore, our framework is able to process a large number of arbitrary websites with no privacy implications.

\noindent\textbf{Defense of password guessing attacks.} In addition to credential stuffing, password guessing has been another major branch of password attack research~\cite{wang2016targeted, thomas2019protecting, pasquini2021improving}. Attackers aim to recover billions of encrypted passwords offline or submit guessable passwords to online login interfaces. To combat these attacks, researchers have proposed password strength meters that guide users to create strong passwords based on statistical prediction models~\cite{dell2015monte, wheeler2016zxcvbn}. Beyond addressing guessable password patterns, researchers have explored the detection of abnormal authentication traffic using traffic fingerprinting techniques~\cite{herley2019distinguishing, tian2020stopguessing}. These solutions have been widely adopted and enhanced password security in general. However, we need completely distinct countermeasures for credential stuffing attacks, which exploit reused passwords regardless of their guessability.

\section{Conclusion}
\jh{
We propose \oursNew{}, a framework for predicting the risk of credential stuffing between websites. We introduce the concept of a password reuse relation---the relationship between websites where users are highly likely to reuse passwords---and represent these relations as edges in a website graph. To predict the existence of such edges, we design a graph neural network (GNN)-based model. Our approach is scalable to a large number of arbitrary websites by extracting public features and incorporating newly observed websites through complementary edges. To preserve privacy, we adopt a federated learning (FL) approach that enables risk prediction without sharing user sensitive information from potentially untrustworthy administrators. Evaluation on a real-world credential breach dataset demonstrates the effectiveness of our model, achieving an F1-score of 0.9153 in the FL setting. Furthermore, we show that our FL-based GNN outperforms other state-of-the-art GNN models, with performance gains of 4-11\%. By employing \oursNew{}, we believe that website administrators can proactively deter credential stuffing attacks by identifying password reuse risks in advance.
}



\printcredits

\bibliographystyle{unsrt}

\bibliography{references}


\clearpage

\bio{bio_jaehan}
\textbf{Jaehan Kim} is a PhD candidate in the School of Electrical Engineering at KAIST. He received his M.S. degree and B.S. degree in School of Electrical Engineering at KAIST. His current research interests mainly focus on AI security, particularly on the safety of large language models (LLMs), and AI-driven solutions for cybersecurity.
\endbio

\vspace{0.5in}

\bio{bio_song}
\textbf{Minkyoo Song} is a PhD candidate in the School of Electrical Engineering at KAIST. He received his M.S. degree in Electrical Engineering at KAIST and his B.S. degree in Industrial Engineering and double majored in Electrical Engineering at KAIST. His current research interests mainly focus on Large language model security, AI security, and Data mining.
\endbio

\vspace{0.5in}

\bio{bio_seo}
\textbf{Minjae Seo} is a Researcher at ETRI, Daejeon, South Korea. He received his M.S. degree from the Graduate School of Information Security at KAIST and his B.S. degree in Computer Engineering from Mississippi State University. His current research interests include network fingerprinting, deep learning-based network system, and AI for security.
\endbio

\vspace{0.5in}

\bio{bio_youngjin}
\textbf{Youngjin Jin} is a PhD candidate in the School of Electrical Engineering at KAIST. He received his M.S. degree in Electrical Engineering at KAIST and his B.S. degree in Electrical Engineering with a minor in Computer Science at KAIST. His current research interests are on large language model security, with a focus on applications on cybersecurity and cyber threat intelligence.
\endbio

\vspace{0.5in}


\bio{bio_shin}
\textbf{Seungwon Shin} is an Associate Professor in the School of Electrical Engineering at KAIST and an Executive Vice President at Samsung Electronics. He received his Ph.D. in Computer Engineering from the Electrical and Computer Engineering Department at Texas A\&M University and his M.S. degree and B.S. degree from KAIST, both in Electrical and Computer Engineering. His research interests span the areas of Software-defined networking security, DarkWeb analysis, and Cyber threat intelligence.
\endbio

\vspace{0.5in}

\bio{bio_jinwoo}
\textbf{Jinwoo Kim} is an Assistant Professor in the School of Software at Kwangwoon University, Seoul, Republic of Korea. He received his Ph.D. from the School of Electrical Engineering at KAIST, his M.S. degree from the Graduate School of Information Security at KAIST, and his B.S. degree from Chungnam National University in Computer Science and Engineering. His research topic focuses on investigating security and privacy issues with software-defined networks, cloud systems, and VR.
\endbio



\end{document}